\documentclass{article}
\usepackage[a4paper, total={6in, 8in}]{geometry}
\usepackage{graphicx}
\usepackage{subcaption}
\usepackage[dvipsnames]{xcolor}
\usepackage[round]{natbib}
\bibpunct{(}{)}{;}{a}{,}{\text{\,\&\,}}
\usepackage{jcappub}
\usepackage[T1]{fontenc} 
\usepackage[most]{tcolorbox}
\usepackage{array}
\usepackage{appendix}
\usepackage{hyperref}
\usepackage{tikz}
\usepackage{adjustbox}
\usepackage[most]{tcolorbox}
\usetikzlibrary{fit, backgrounds, shapes.misc, shapes.geometric, arrows.meta, positioning, calc}

\title{Competing with AI Scientists: Agent-Driven Approach to Astrophysics Research}

\author[1,2,*]{Thomas Borrett}
\author[2, 3, *]{Licong Xu}
\author[1,2, *]{Andy Nilipour}
\author[1,2]{Boris Bolliet}
\author[4]{Sebastien Pierre}
\author[4]{Erwan Allys}
\author[1,5]{Celia Lecat}
\author[6]{Biwei Dai}
\author[7]{Po-Wen Chang}
\author[7]{Wahid Bhimji}
\emailAdd{tb711@cam.ac.uk}
\emailAdd{an755@cam.ac.uk}
\emailAdd{lx256@cam.ac.uk}

\affiliation[1]{Cavendish Astrophysics, University of Cambridge, Madingley Road, Cambridge CB3 0HA, UK}
\affiliation[2]{Kavli Institute for Cosmology, University of Cambridge, Madingley Road, Cambridge CB3 0HA, UK}
\affiliation[3]{Institute of Astronomy, University of Cambridge, Madingley Road, Cambridge CB3 0HA, UK}
\affiliation[4]{Laboratoire de Physique de l'\'{E}cole normale sup\'erieure, ENS, Universit\'e PSL, CNRS, Sorbonne Universit\'e, Universit\'e Paris Cit\'e, F-75005 Paris, France}
\affiliation[5]{CentraleSupélec, Gif-sur-Yvette, France}
\affiliation[6]{School of Natural Sciences, Institute for Advanced Study, 1 Einstein Drive, Princeton, NJ 08540, USA}
\affiliation[7]{NERSC, Lawrence Berkeley National Laboratory, Berkeley, CA 94720, USA}
\affiliation[*]{Equal Contribution}

\date{\today}

\abstract{
    We present an agent-driven approach to the construction of parameter inference pipelines for scientific data analysis. Our method leverages a multi-agent system, \href{https://github.com/CMBAgents/cmbagent}{\texttt{Cmbagent}}\footnote{\url{https://github.com/CMBAgents/cmbagent}} (the analysis system of the AI scientist \href{https://github.com/AstroPilot-AI/Denario}{\texttt{Denario}}\footnote{\url{https://github.com/AstroPilot-AI/Denario}}), in which specialized agents collaborate to generate research ideas, write and execute code, evaluate results, and iteratively refine the overall pipeline. As a case study, we apply this approach to the FAIR Universe Weak Lensing Uncertainty Challenge, a competition under time constraints focused on robust cosmological parameter inference with realistic observational uncertainties. While the fully autonomous exploration initially did not reach expert-level performance, the integration of human intervention enabled our agent-driven workflow to achieve a first-place result in the challenge. This demonstrates that semi-autonomous agentic systems can compete with, and in some cases surpass, expert solutions. We describe our workflow in detail, including both the autonomous and semi-autonomous exploration by \href{https://github.com/CMBAgents/cmbagent}{\texttt{Cmbagent}}. Our final inference pipeline utilizes parameter-efficient convolutional neural networks, likelihood calibration over a known parameter grid, and multiple regularization techniques. Our results suggest that agent-driven research workflows can provide a scalable framework to rapidly explore and construct pipelines for inference problems. 
}

\begin{document}

\maketitle

\tableofcontents

\section{Introduction}

Modern scientific inference requires navigating increasingly high-dimensional \emph{methodological design spaces} spanning data representations and augmentations, model architectures, optimization strategies. Exhaustive human exploration of these spaces is often intractable. Recent advances in large language models (LLMs) enable \emph{agentic} systems that can propose methods, write and run code, analyze outcomes, and iterate via feedback loops. These capabilities suggest a new paradigm in which autonomous agents assist researchers in navigating complex problems to accelerate scientific discovery.

In this article, we present a case study of \emph{agent-driven discovery} for cosmological parameter inference with weak gravitational lensing. Our primary contribution is methodological: we demonstrate how a multi-agent research workflow, when combined with human interactions, can refine analysis strategies and produce pipelines competitive with state-of-the-art approaches. While the physics behind weak gravitational lensing is well understood, the inverse problem is complicated by limited training data, non-Gaussian information that classical methods fail to utilize, and the need for well-calibrated uncertainties.

Agentic systems are gaining popularity within cosmology \citep{moss2025aicosmologistiagentic, peng2026deepinflationaiagentresearch, mudur2025llm, sun2025mephistoselfimprovinglargelanguage, ting2025egentautonomousagentequivalent, ramachandra2025teachingllmsspeakspectroscopy, li2026madevolveevolutionaryoptimizationcosmological, casas2025clappclassllmagent} and science more generally \citep{denario, ghafarollahi2024sciagentsautomatingscientificdiscovery,  wei2025, pinheiro2025largelanguagemodelsachieve, gao2025testtimescalingtechniquestheoretical, nagele2026agenticexplorationphysicsmodels, ye2025replicationbenchaiagentsreplicate, zhang2025bridgingliteratureuniversemultiagent, arlt2025autonomousquantumphysicsresearch, miao2025physmasterbuildingautonomousai}. We employ \href{https://github.com/CMBAgents/Cmbagent}{\texttt{Cmbagent}}\citep{Laverick:2024fyh,Cmbagent_2025,xu2025opensourceplanning}, which uses \href{https://github.com/ag2ai/ag2}{\texttt{ag2}}\footnote{\url{https://github.com/ag2ai/ag2}} as a scaffold, as our multi-agent system to investigate the space of modeling and inference choices.

Using this framework, we systematically explore architectures, training strategies, and inference methods. We find that fully autonomous exploration of the design space often focuses on inefficient and performance-saturated methods, whereas interfacing with humans (i.e.,\,\emph{human-in-the loop} intervention) results in more focused and insightful design choices. The final pipeline combines a multi-scale Inception-style convolutional neural network \citep[CNN;\,][]{szegedy2014goingdeeperconvolutions}, symmetry-aware data transformations, and likelihood calibration to produce accurate cosmological parameter estimates with robust uncertainty quantification. The pipeline assembles components from prior work in cosmological inference \citep{Ribli_2019, PhysRevD.110.043535}, extends them to non-square maps, and incorporates regularization methods for likelihood estimation from finance \citep{LedoitWolf2003}.  While the application in this work is to weak lensing, our framework is general: agentic workflows provide a practical mechanism for systematically exploring complex methodological spaces in scientific inference and for rapidly assembling accurate, competitive pipelines. 

The rest of the article is organized as follows. In Section \ref{sec:agents}, we demonstrate a human-agent research loop that can identify uncommon but effective architectural and inference choices. In Section \ref{sec:methods}, we present our search procedure and the final pipeline resulting from human-agent discovery, and we show that it is competitive with state-of-the-art inference in Section \ref{sec:results}. In \ref{sec:discussion}, we discuss insights into agent-driven workflows and analyze failure modes of fully autonomous discovery, and we conclude in Section \ref{sec:conclusion}.


\section{Agent-Driven Method Discovery}
\label{sec:agents}

\begin{figure}[!ht]
\centering
\begin{tikzpicture}[
    node distance=3.8cm,
    label_style/.style={font=\footnotesize\sffamily\bfseries, color=gray!80!black},
    arrow/.style={-{Stealth[scale=1.2]}, line width=3pt, color=gray!80}
]

\node (map) [inner sep=0pt] {
    \includegraphics[height=4cm]{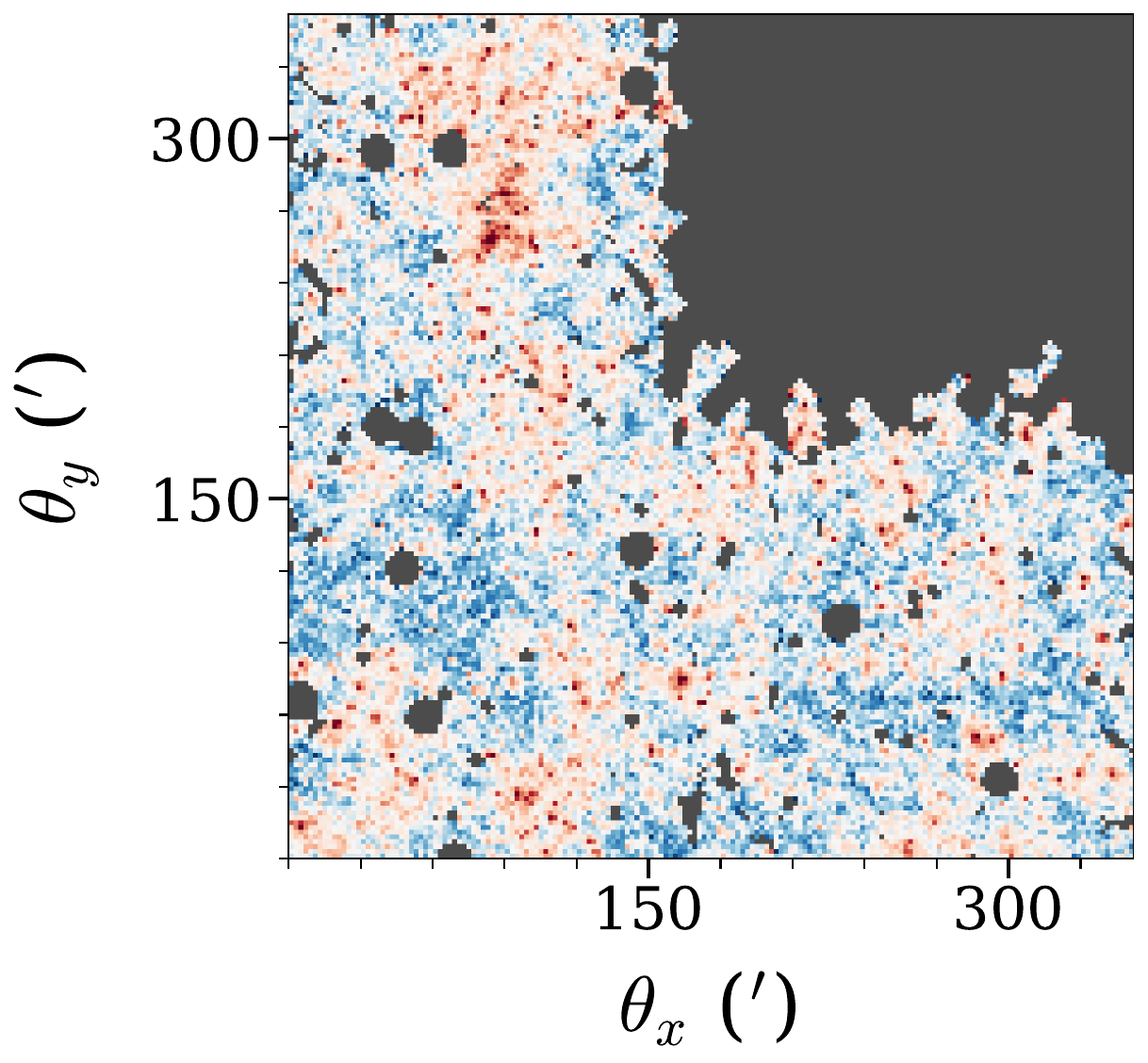}
};
\node[label_style, below=0.1cm of map] {Input: Map $x$};


\node (posterior) [right=of map, inner sep=0pt] {
    \includegraphics[height=4cm]{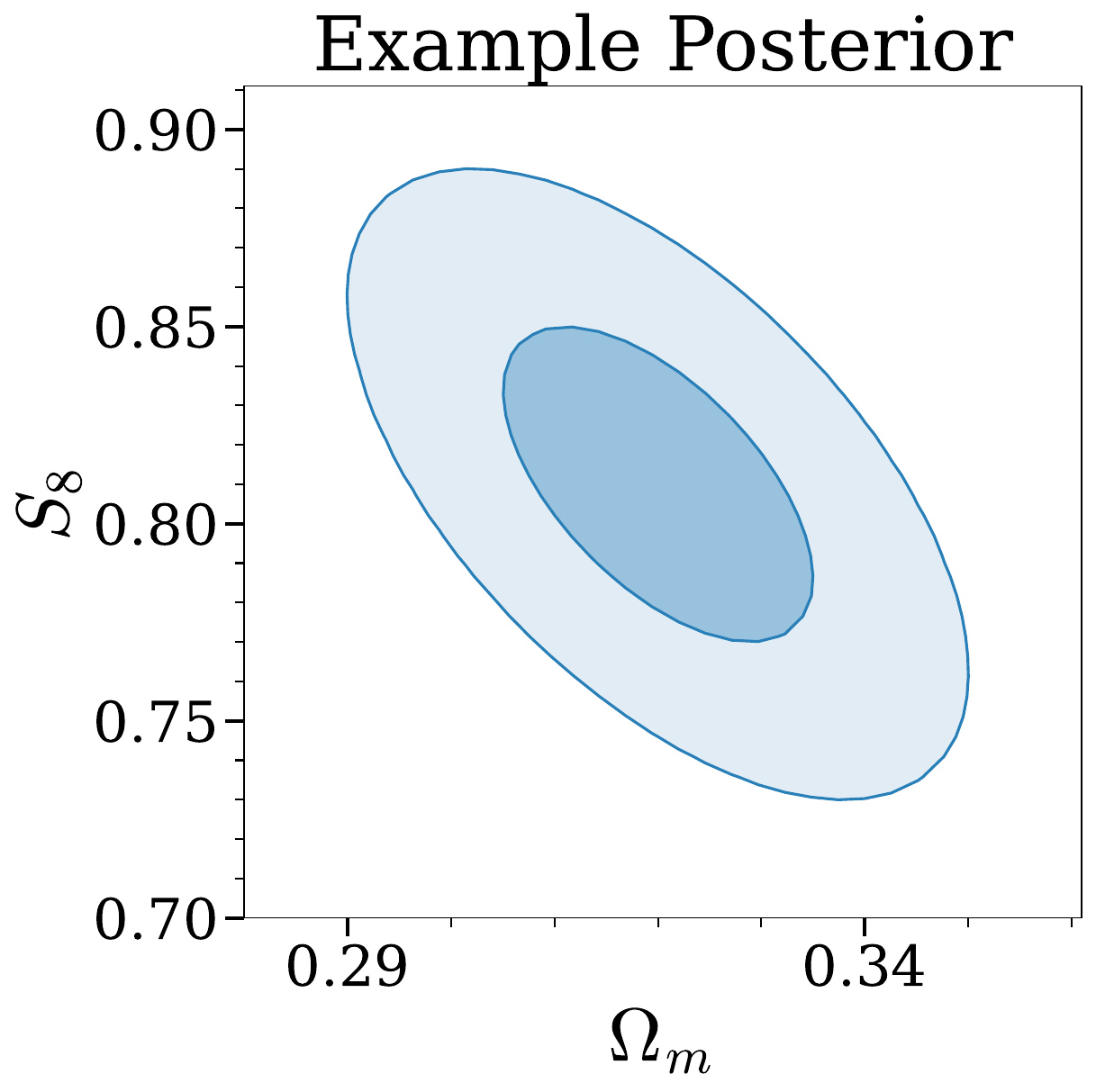}
};
\node[label_style, below=0.1cm of posterior] {Output: $P(\theta|x)$};

\draw [arrow] (map.east) -- (posterior.west) 
    node[midway, above=0.3cm, align=center, font=\sffamily\color{black}, color=black] {
        \textbf{Inference Pipeline} \\
        \small $f: x \mapsto (\hat{\theta}, \hat{\sigma})$
    };


\end{tikzpicture}
\caption{Inference pipeline for weak lensing. The objective is to develop a mapping, $f$, that transforms an input weak lensing map, $x$, into predicted cosmological parameters, $\hat{\theta}$, and associated uncertainties, $\hat{\sigma}$. This process yields a joint posterior distribution (right) for the cosmological parameters $\Omega_m$, the matter density, and $S_8$, the clustering amplitude. The input map shown is a square 176 pixel-wide cutout from the full map, which has pixel dimension $1424\times 176$.}
\label{fig:inference-example}
\end{figure}
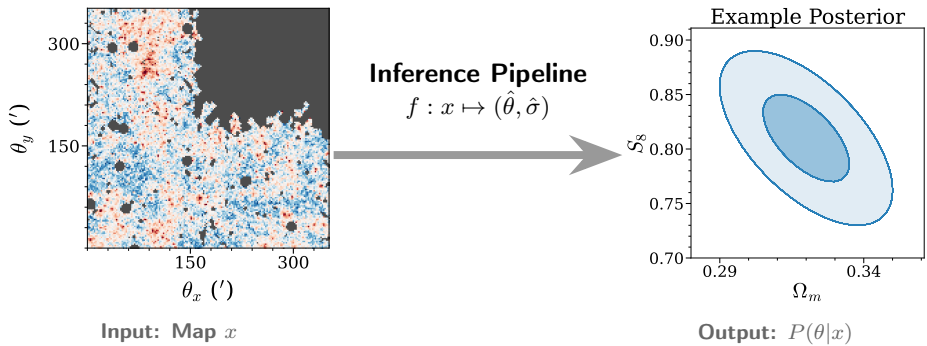

To make the methodology concrete, we briefly formalize the structure of the multi-agent system used in this work. We leverage an agent-driven framework that employs diverse LLMs, orchestrated dynamically through reasoning and tool use to enable effective agentic collaboration. A single agent, denoted by index $i$, can be formulated as $a_i = (L_i, P_i, T_i)$, where $L_i$ is the LLM model, $P_i$ is the context information (e.g.,\,system prompt and context variables), and $T_i$ is the set of available tools. A multi-agent system is then defined as the tuple $\mathcal{M} = (\mathcal{A}, \Phi, \mathcal{H})$, where $\mathcal{A}=\{a_1,...,a_n\}$ is a collection of $n$ agents, $\Phi$ is the orchestration function that manages multi-agent collaboration, and $\mathcal{H}$ denotes the human intervention layer for supervision and feedback.

In the context of \texttt{Cmbagent}, the agentic framework is instantiated into distinct configurations of $\mathcal{M}$ depending on the operational mode. The system provides a \emph{One-Shot} mode, in which the task is directly passed to a user-specified agent $a_i$ (e.g.,\,the \texttt{engineer} for code generation). In this case, the orchestration function $\Phi$ represents an iterative loop between code generation and execution until the termination condition is met. 

Alternatively, the \emph{Planning \& Control} mode involves higher levels of orchestration, where $\Phi$ is decomposed into a sequence of a \emph{planning} phase and a \emph{control} phase, $\Phi = (\Phi_\text{plan}, \Phi_\text{control})$. In this configuration, $\Phi_\text{plan}$ involves a \emph{planner-reviewer} loop to decompose the complex research task into smaller subtasks. Subsequently, $\Phi_\text{control}$ manages the transitions between the downstream specialized agents $a_s \subset \mathcal{A}$ to execute those subtasks. In both modes, there is no human-in-the-loop beyond the initial user request prompt, resulting in a fully autonomous execution where $\mathcal{H}=\emptyset$. We refer the reader to \cite{xu2025opensourceplanning} for details on the architectures and orchestrations of \texttt{Cmbagent}.

Having outlined the agentic framework, we now describe the scientific task used to evaluate it. To demonstrate how LLMs and agentic systems can be applied to research, we use weak gravitational lensing as a representative case of a general inference problem. While we focus here on an astrophysical application, our findings and the underlying framework remain relevant to any scientific domain. The work presented in this paper was completed to compete in the FAIR Universe Weak Lensing Uncertainty Challenge\footnote{\url{https://www.codabench.org/competitions/8934/}}, with the objective of obtaining point estimates and calibrated uncertainties from realistic weak lensing maps. More details and results of the challenge will be presented in a forthcoming paper by the FAIR Universe organizers and other participating teams.

\section{Methods} \label{sec:methods}
\subsection{Data}
Physically, weak gravitational lensing is the deflection of light from background galaxies by density perturbations along the line of sight. Statistical measurements of the distortion of these background galaxies can probe the inhomogeneous distribution of matter in the Universe, enabling the inference of cosmological parameters. For a review of the physics and methodology, see \citet{Kilbinger_2015}. While the lensing is approximately Gaussian at large scales, where it can be characterized by two-point statistics such as the power spectrum, precise parameter inference requires extracting useful information contained within the non-Gaussian features in the cosmic web.

The competition dataset consists of simulated observations from the second redshift bin of the Hyper Suprime-Cam (HSC) survey \citep{hsc_ssp, hsc_y3}. One section of a sample map is shown in the left panel of Figure \ref{fig:inference-example}. Each map is labeled with two cosmological parameters: the matter density of the universe, $\Omega_m$, and the fluctuation amplitude, $S_8 = \sigma_8 \sqrt{\Omega_m/0.3}$, where $\sigma_8$ is the amplitude of matter density fluctuations on scales of $8 \ \text{Mpc} \ h^{-1}$. To capture the complexities of modern cosmological surveys, the simulations include three systematic nuisance parameters: the two baryonic parameters are the active galactic nuclei (AGN) temperature, $\log_{10}(T_\text{AGN}/K)$, and the present stellar mass fraction, $f_0$, while the third nuisance parameter, $\Delta z$, is the photometric redshift error. The maps provided in the training set are noiseless, but the test maps include pixel-level noise characterized by a shape noise level of $\sigma_\epsilon=0.4$ with a galaxy density $n_g=30$ $\text{arcmin}^{-2}$. We note that this task operates in a low-data regime under simplified assumptions (e.g.,\,a single redshift bin and no intrinsic alignment). The training dataset comprises 101 distinct pairs of $(\Omega_m, S_8)$, with 256 different nuisance parameter realizations for each cosmology. We split this into a training and validation dataset.

\subsection{Inference Task and Evaluation Objective} \label{sec:prob_eval}

We now formalize the inference task. Let $x \in \mathcal{X}$ denote a weak lensing map generated from a simulation with cosmological parameters $\theta = \left(\Omega_m, S_8\right) \in \Theta$. We aim to construct an inference pipeline, 
\begin{align}
    f : \mathcal{X} \rightarrow \hat{\Theta},\ f(x) =(\hat{\theta}, \hat{\sigma}),
\end{align} 
which maps each realization $x$ to a point estimate, $\hat{\theta} = \left(\hat{\Omega}_m, \hat{S}_8\right)$, and its associated one-standard deviation uncertainties, $\hat{\sigma} = \left(\hat{\sigma}_{\Omega_m}, \hat{\sigma}_{S_8}\right)$. Performance is evaluated using a scalar scoring function, $s(f(x), \theta)$, that jointly assesses the accuracy of the point estimate and the calibration of the predicted uncertainties,
\begin{align}
\begin{split}
s(f(x), \theta) = & - \left\{\frac{(\hat{\Omega}_{m} - \Omega_{m})^2}{\hat{\sigma}_{\Omega_{m}}^2} + \frac{(\hat{S}_{8} - S_{8})^2}{\hat{\sigma}_{S_{8}}^2} + \log (\hat{\sigma}_{\Omega_{m}}^2) + \log (\hat{\sigma}_{S_{8}}^2)\right. \\
& \left.  + \lambda \left[(\hat{\Omega}_{m} - \Omega_{m})^2 + (\hat{S}_{8} - S_{8})^2 \right] \right\}.
\end{split}
\label{eq:score}
\end{align}
This score is a modified Kullback-Leibler (KL) divergence, where the first four terms represent a Gaussian log-likelihood, penalizing both bias and miscalibration. The final term is a mean-squared error (MSE) loss with $\lambda=10^3$ to penalize poor point estimates. The inference task is an optimization problem to find $f^*$ that maximizes the average score over the set $\mathcal{X}$ of test data:
\begin{align}
    f^* = \underset{f\in\mathcal{F}}{\arg \max} \ \frac{1}{|\mathcal{X}|}\sum_{x\in\mathcal{X}}s(f(x), \theta),
\end{align}
where $\mathcal{F}$ denotes the space of inference pipelines. This objective defines the search space to be explored. 



\subsection{Autonomous Agent discovery}

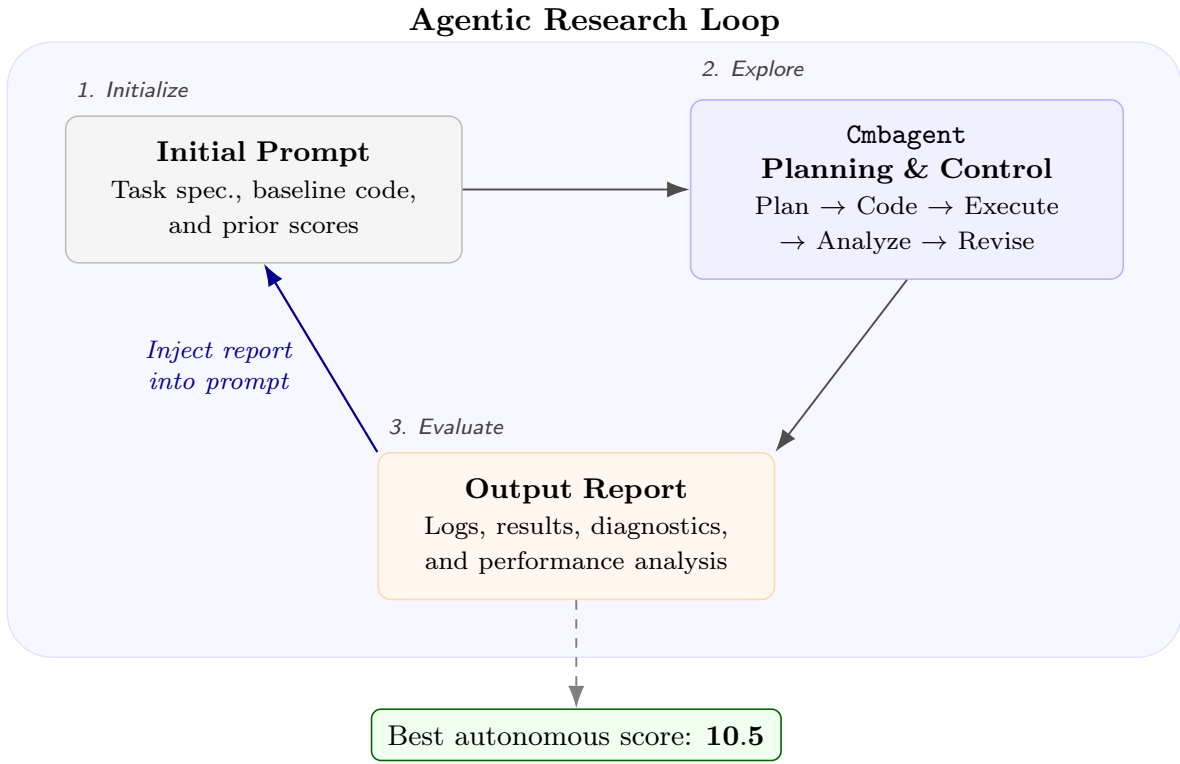
\begin{figure}[t]
\centering
\resizebox{\linewidth}{!}{%
\begin{tikzpicture}[
    font=\small,
    box/.style={
        draw, rounded corners=4pt, align=center,
        inner sep=8pt, line width=0.5pt, text width=3.8cm,
        minimum height=1.4cm
    },
    prompt/.style={box, fill=gray!8, draw=gray!50},
    workflow/.style={box, fill=blue!6, draw=blue!30, text width=4.2cm},
    report/.style={box, fill=orange!6, draw=orange!30},
    score/.style={
        draw=green!40!black, rounded corners=3pt, fill=green!6,
        align=center, inner sep=5pt, line width=0.5pt,
        font=\small
    },
    arr/.style={-{Latex[length=2.5mm, width=1.8mm]}, line width=0.6pt, color=gray!60!black},
    feedarr/.style={-{Latex[length=2.5mm, width=1.8mm]}, line width=0.7pt, color=blue!55!black},
    steplabel/.style={font=\scriptsize\sffamily, text=gray!50!black},
]


    \node[prompt] (init) {%
        \textbf{Initial Prompt}\\[2pt]
        {\footnotesize Task spec., baseline code,\\and prior scores}%
    };

    \node[workflow, right=2.5cm of init] (workflow) {%
        \texttt{Cmbagent}\\
        \textbf{Planning \& Control}\\[2pt]
        {\footnotesize Plan $\to$ Code $\to$ Execute\\$\to$ Analyze $\to$ Revise}%
    };

    \node[report, below=2.0cm of $(init.south east)!0.5!(workflow.south west)$] (report) {%
        \textbf{Output Report}\\[2pt]
        {\footnotesize Logs, results, diagnostics,\\and performance analysis}%
    };

    \node[score, below=1.2cm of report] (scorebox) {%
        Best autonomous score: $\mathbf{10.5}$%
    };

    \node[steplabel, above=0.08cm of init.north west, anchor=south west] {\textit{1. Initialize}};
    \node[steplabel, above=0.08cm of workflow.north west, anchor=south west] {\textit{2. Explore}};
    \node[steplabel, above=0.08cm of report.north west, anchor=south west] {\textit{3. Evaluate}};


    \draw[arr] (init.east) -- (workflow.west);

    \draw[arr] (workflow.south) -- (report.north east);

    \draw[feedarr] (report.north west) -- (init.south)
        node[pos=0.45, left=0.25cm, text=blue!55!black, font=\footnotesize\itshape, align=center]
        {Inject report\\into prompt};

    \draw[arr, dashed, gray!100] (report.south) -- (scorebox.north);

    \begin{scope}[on background layer]
        \node[
            fill=blue!3,
            draw=blue!12,
            line width=0.4pt,
            rounded corners=14pt,
            fit=(init) (workflow) (report),
            inner sep=18pt,
            label={[font=\normalsize, text=black, anchor=south, yshift=-2pt]above:\textbf{Agentic Research Loop}}
        ] (bg) {};
    \end{scope}

\end{tikzpicture}%
}
\caption{The autonomous research loop we adopt for this task. We craft an initial prompt with example code, context, and scores of various approaches. \texttt{Cmbagent} in \emph{Planning \& Control} mode then attempts a solution according to a plan, with analysis of errors and shortcomings. The output report is injected into the initial prompt, and the loop is restarted for a second attempt.}
\label{fig:autonomous-cmbagent}
\end{figure}

We now describe how the agentic system explores this space in practice. The optimization problem defined in Section \ref{sec:prob_eval} involves a high-dimensional search over model architectures and training techniques. Significant manual effort is required to explore this space $\mathcal{F}$ and to identify optimal configurations, a process that is often inefficient and prone to human bias. We address this challenge by employing agentic systems with two objectives: first, to accelerate the discovery process through autonomous and effective iteration, and second, to evaluate the capability of LLMs in navigating complex physical domains where optimal strategies may diverge from established methods by human experts in the field.


To perform the search for $f^* \in \mathcal{F}$, we first use the \emph{Planning \& Control} mode of \texttt{Cmbagent} with no human-in-the-loop ($\mathcal{H}=\emptyset$). The search is initialized by a structured user prompt on task specification, $P_\text{user}$, which serves as the input to the \textit{planner} agent, $a_\text{planner}=(L_\text{planner}, P_\text{system}+P_\text{user},\emptyset)$. Here, $L_\text{planner}=\texttt{GPT-4.1}$, $P_\text{system}$ is the system prompt, and the planner operates without external tool calls. This prompt induces in-context learning by encoding empirically grounded priors, including baseline insights and historical performance benchmarks. The orchestration begins after the user inputs the prompt. Figure \ref{fig:autonomous-cmbagent} illustrates this autonomous discovery process. For an example prompt, see Appendix \ref{app:prompt}.

While this design enables the rapid identification of candidate pipelines, the search trajectory often reveals specific failure modes. \texttt{Cmbagent} in \textit{Planning \& Control} mode struggles to qualitatively assess its implementations. We return in Section \ref{sec:results} to show how this setup performs and compares to other integrations between human researchers and LLMs.

\subsection{Human-Agent Co-discovery}

Our second integration mode adopts a hybrid approach, with human-in-the-loop interventions to guide exploratory progress and validate outputs. Rather than operating autonomously, the generation of hypotheses, analyses, and solution candidates by agents was shaped through iterative interaction. We attempt this in two ways. Firstly, we monitor code running in the \emph{Planning \& Control} mode to qualitatively assess if it is a useful direction, and we terminate the run early if it is not ($\mathcal{H}\neq \emptyset$). Secondly, we utilize the \emph{One-Shot} mode of \texttt{Cmbagent} using our own expertise and subjective judgment to alter the prompt and impose further empirically grounded priors for the planner. This arises through interaction with the plan via $P_\text{user}$ and is analogous to the research loop presented in Figure \ref{fig:autonomous-cmbagent}, with the key modification that we collect the output report and alter the initial prompt. Note that \textit{One-Shot} mode only performs a single pass with code written, executed and debugged automatically. In practice, this intervention involves stopping the focus on CNN architectures that exhibit diminishing returns by imposing simplicity or parameter-efficiency constraints. Feedback is also used to encourage discovery of diverse strategies, rather than continual refinement of a single pipeline.

\subsection{Final Inference Architecture} \label{sec:inference}

Guided by the exploration strategies described above, we now present the final inference pipeline that emerged. The choices were driven by iterative exploration, and the constraints of various approaches were revealed during the search process. The low-data regime meant that parameter-efficient approaches used in conjunction with ensembles of CNN models proved more accurate than a single high-capacity model. Incorporation of data augmentation was extremely important, during both training and inference. 

\begin{figure}[t]
\centering
\resizebox{0.8\linewidth}{!}{%
\begin{tikzpicture}[
    font=\small,
    node distance=10mm and 12mm,
    box/.style={draw, rounded corners, align=center, inner sep=6pt, line width=0.5pt, text width=40mm},
    data/.style={box, fill=gray!5},
    proc/.style={box, fill=blue!5},
    stat/.style={box, fill=orange!5},
    final/.style={box, fill=green!5, line width=1pt},
    arr/.style={-Latex, line width=0.7pt},
    darr/.style={-Latex, line width=0.7pt, dashed},
]

\node[data] (x) {\textbf{Convergence Map}\\$x \in \mathcal{X}$};
\node[proc, below=of x] (cnn) {\textbf{CNN + SC}\\ CNN + SC predicts $\hat\theta (g(x))$};
\node[proc, below=of cnn] (agg) {\textbf{Average}\\Avg. over $D_4$ symmetries};

\node[data, right=15mm of x] (val) {\textbf{Validation Set}\\Maps with known $\theta_g$};
\node[stat, below=of val, right=15mm of cnn] (cal) {\textbf{Likelihood Calibration}\\Collect $(\hat\theta, \theta_g)$ pairs\\Estimate $\mu_g, \Sigma_g$ at each point $g$};
\node[stat, below=of cal, right=15mm of agg] (like) {\textbf{Likelihood Model}\\$p(\hat\theta \mid \theta_g) \approx \mathcal{N}(\hat\theta; \mu_g, \Sigma_g)$ for predicted $\hat\theta$ from model};


\node[stat, below=of like] (post) {\textbf{Posterior Weights}\\$w_g \propto p(\hat{\theta} | \theta_g)$ over the cosmology points};
\node[final, left=of post, text width=25mm] (res) {\textbf{Outputs}\\$\hat\theta$, $\hat\sigma$};

\draw[arr] (x) -- (cnn);
\draw[arr] (cnn) -- (agg);
\draw[arr] (agg) -- (like); 
\draw[arr] (val) -- (cal);
\draw[arr] (cal) -- (like);
\draw[arr] (like) -- (post);
\draw[arr] (post) -- (res);
\path (agg.east) -- (like.west) coordinate[midway] (mid);
\draw[arr] (agg.east) -- (mid) |- (cal.west);

\begin{scope}[on background layer]
    \node[fill=blue!10, rounded corners, fit=(x) (cnn) (agg), label=above:\textbf{Feature Extraction}] {};
    \node[fill=orange!10, rounded corners, fit=(val) (cal) (like) (post), label=above:\textbf{Inference}] {};
\end{scope}

\end{tikzpicture}%
}
\caption{Diagram of our inference pipeline. CNN ensemble predictions (with $D_4$ test-time augmentation) are treated as summary statistics, $\hat\theta$; an empirical Gaussian likelihood is calibrated on validation predictions at each cosmology point $\theta_g$ and is used to estimate the weights to produce point estimates and uncertainties.}
\label{fig:inference-pipeline}
\end{figure}
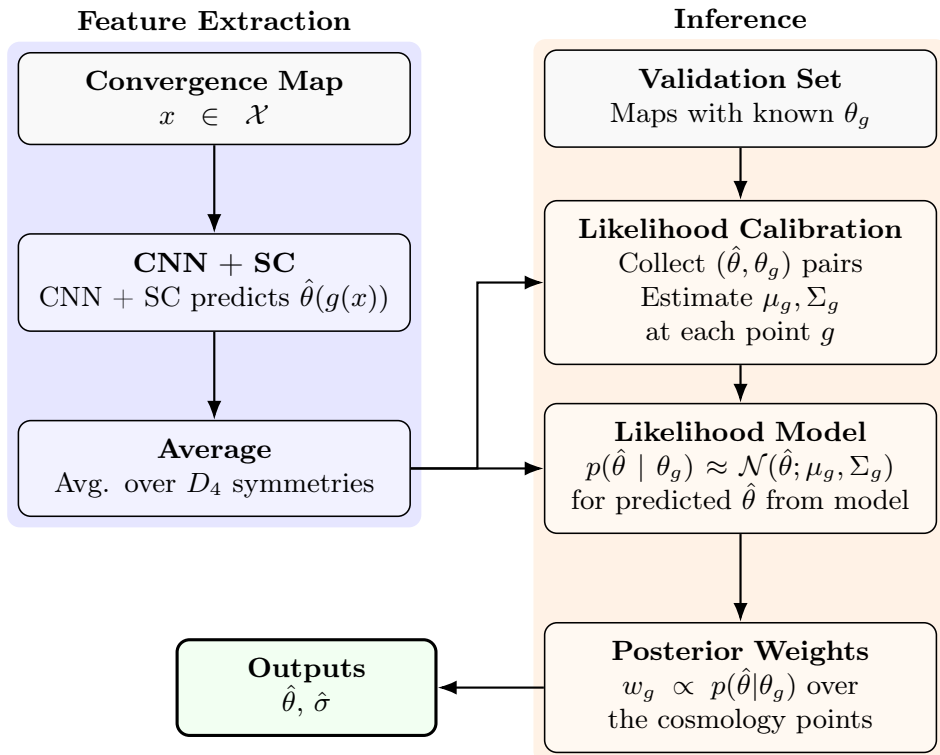

The best-performing pipeline separates inference and regression, and it treats the output of CNNs as a summary statistic. Uncertainty quantification uses a calibrated likelihood on the distribution of validation predictions by an ensemble of CNN models. Architectural choices for the CNN led to small models which are capable of multi-scale feature extraction. Our final pipeline uses Inception and Inception Squeeze-and-Excitation (InceptionSE) networks, which are ideal for this application because of their varying kernel size. For full details on the CNN architectures, we refer to Appendix \ref{app:inceptionse}. A diagram of our inference pipeline is shown in Figure \ref{fig:inference-pipeline}. 

\subsubsection{CNN Predictions as Summary Statistics}

We found the raw CNN predictions to be systematically biased, and we therefore treated them as summary statistics rather than final parameter estimates. To account for this, we augment the predictions with a small correction term based on scattering covariances \citep[SC;\,][]{Cheng2024} and perform a separate likelihood-based inference step to recover calibrated parameter estimates and uncertainties. The main stages of this procedure are summarized in Algorithm 1; here we describe the core likelihood construction.

Our trained CNNs map each convergence map $x$ to a two-dimensional prediction, $\hat{\theta}$. For each ensemble member $m$, we collect prediction--truth pairs on that member's held-out validation maps. Grouping these predictions by their true cosmology allows us to empirically estimate, at each cosmology point $\theta_g$, a predicted mean $\mu_g$ and covariance $\Sigma_g$. We then define a Gaussian likelihood model for a test-time prediction $\hat{\theta}$:
\begin{align}
    p(\hat{\theta} \mid \theta_g) \simeq \mathcal{N}(\hat{\theta}; \mu_g, \tilde{\Sigma}_g)\, .
\end{align}
An important component of the final pipeline is test-time augmentation using all symmetries of the dihedral group $D_4$, followed by averaging the corresponding predictions in parameter space.

Given the likelihood calibrated over grid points, we form a discrete posterior over $\theta_g$ and compute the posterior mean and marginal variances as
\begin{align}
    \hat{\theta}_\text{post} &= \sum_g w_g \theta_g\, , \\
    \sigma^2_\text{post} &= \sum_g w_g (\theta_g - \hat{\theta}_{\text{post}})^2 \, .
\end{align}
Further details of the inference pipeline, including likelihood calibration and hyperparameter optimization, are provided in Appendix \ref{app:inference-details}.

\section{Results}
\label{sec:results}

We now evaluate the inference pipeline obtained through the agent-driven discovery described above. Performance is assessed on simulated weak lensing convergence maps using the scoring function described in Equation \ref{eq:score} and compared to other competitors who took part in the FAIR Universe challenge, which we use as a baseline. We achieved a first place result 
in both the public phase, and in one of three final \href{https://fair-universe.lbl.gov/WeakLensing-Uncertainty-Challenge.html}{leaderboards}\footnote{\url{https://fair-universe.lbl.gov/WeakLensing-Uncertainty-Challenge.html}}. This assessed the pipeline on unseen simulations but over the same cosmologies as in the training set. Further details are available on the competition webpage and will be discussed in the forthcoming NeurIPS 2026 paper. All reported results are from this leaderboard, and we provide both point-estimate accuracy and uncertainty calibration through coverage.

\begin{table}[!ht]
\centering
\begin{tabular}{l c c c}
\hline
& \textbf{Score} & \textbf{MSE} & \textbf{Coverage} \\
\hline
 & 11.70 & 0.10 & 0.70 \\
\hline
\end{tabular}
\caption{Performance of the inference pipeline on an unseen test set. The MSE is the normalized error, and the coverage is the fraction of parameters within $\pm1\sigma$.}
\label{tbl:final_results}
\end{table}

Our final score of 11.70 (Table \ref{tbl:final_results}) achieved first place in FAIR Universe Weak Lensing Uncertainty Challenge, outperforming all other submitted approaches, including pipelines developed by domain experts in weak lensing and machine learning \citep{LSST2026}. In addition to achieving the top overall score, the method attains a mean squared error of 0.10 and a coverage of 0.70, indicating both accurate point estimates and well-calibrated uncertainty. This result demonstrates that, in a well-defined benchmarking setting, integrating agentic systems into scientific workflows can enable competitive and, in some cases, state-of-the-art performance in complex cosmological inference pipelines.

\begin{table}[!h]
    \centering
    \begin{tabular}{p{3.0cm} p{5.2cm} p{5.6cm}}
        \hline
        \textbf{Category} & \textbf{First surfaced in autonomous search} & \textbf{First surfaced after human redirection} \\
        \hline
        CNN backbone & ResNet (18/34/50) & BaseCNN, Inception, InceptionSE \\
        Inference method & MCMC, direct moments prediction & Grid likelihood over known cosmology points \\
        Augmentation & Flips, rotations & Full $D_4$ symmetries with test-time averaging \\
        \hline
    \end{tabular}
    \caption{Exploration of the inference design space under autonomous and human-redirected agentic search. Entries indicate when a method family first appeared in the search trajectory, rather than a controlled attribution of individual contributions. Some explored methods do not appear in the final pipeline.}
    \label{tab:discovery_trajectory}
\end{table}

This performance should be interpreted in the context of the benchmark structure. The optimized score reflects cosmology labels represented during training. When evaluated on unseen cosmologies, the pipeline performs well in regions with dense nearby training coverage, but degrades substantially in sparse regions of parameter space. This suggests the current method is strong as an on-grid or near-grid posterior estimator, but is not yet a uniformly reliable off-grid inference method. For further discussion, see Appendix \ref{app:param-space}.

\subsection{Failure modes of autonomous discovery} \label{sec:failure_modes}

\begin{table}[!ht]
\centering
\begin{tabular}{l c c c c}
\hline
\textbf{Agents/LLMs} & \textbf{Model} & \textbf{Inference} &
\textbf{Augmentation} & \textbf{Score} \\
\hline
\texttt{Cmbagent}-\emph{P\&C} & ResNet-18/34/50 & MCMC & Flips/rotations & $10.50$ \\
+ \texttt{Cmbagent}-\emph{One-Shot} & BaseCNN & MCMC & No & 11.02 \\
+ \texttt{GPT-5} & BaseCNN & Grid & No & 11.42 \\
+ \texttt{Gemini-2.5-Pro} & Inception + InceptionSE & Grid & Flips/rotations/transpose & 11.72 \\
\hline
\end{tabular}
\caption{Summary of model–generation pathways used during the architecture search, showing the models proposed under different \texttt{Cmbagent}/LLM configurations and their corresponding test-set performance. The ``+'' prefix indicates the incremental integration of the specified LLM into the search pipeline. In the inference method column, “MCMC’’ refers to sampling the posterior using a Markov Chain Monte Carlo algorithm. “Grid’’ correspond to our final inference pipeline, described in Section \ref{sec:inference}. \texttt{BaseCNN} denotes a light-weight CNN. The score presented here is on the validation dataset.
}
\label{tbl:llm_results}
\end{table}

To better understand the role of human intervention, we analyze the performance of fully autonomous configurations. Our results highlight a clear shortcoming of autonomous discovery: our best score was obtained with human intervention inside the agentic loop. This intervention primarily acted to redirect or terminate unproductive search trajectories rather than to propose solutions, while agents generated all candidate architectures and inference strategies. \texttt{Cmbagent} without guidance settles on ResNet \citep{he2016} architectures, which are characterized by their significant depth and high parameter count, and it frequently suggests the next larger iteration without qualitatively exploring alternative parameter-efficient and ultimately better models. Only with human intervention and guidance does it refocus, and through this, we achieve expert-level performance. Table \ref{tbl:llm_results} shows the score progression integrating various LLMs and human-in-the-loop interventions. We note that the competition setting imposed strict time constraints that precluded systematic ablation studies. The progression shown reflects the development of the pipeline rather than a controlled experimental comparison.

\section{Discussion}
\label{sec:discussion}

This case study of applying multi-agent systems to a typical inverse problem yields three key insights: (i) concrete scientific lessons for parameter inference in low-data regimes, (ii) evidence that agent-driven workflows can systematically explore methodological design space beyond standard practice, and (iii) clear limits on autonomous agentic systems in complex scientific problems.

\subsection{Scientific insights}

The ideas presented here were generated and formulated using LLMs and unknown to us before attempting this competition, but well understood and often utilized by the wider machine learning community. 

\paragraph{Ensembles of smaller CNNs achieve competitive performance.} A combination of CNNs with $O(10^6)$ parameters outperform larger, deeper networks, such as ResNet models with $O(10^7)$ parameters, and have a much lower cost to train. This is likely a symptom of the low-data regime we are in, where training models on different training-validation data splits result in more robust inference and a reduced chance of overfitting. As shown in Table \ref{tbl:llm_results}, a very simple \texttt{BaseCNN} model scores higher than 50-layer ResNet models, with $2.5 \times 10^7$ parameters, by +0.5. 

\paragraph{Inception models are particularly efficient for multi-scale feature extraction.} Cosmological inference requires the extraction of multi-scale information from both small-scale features, such as galaxy clustering, and the large-scale cosmic web. This makes Inception-like CNNs, which are parameter efficient and have multiple kernel sizes, perfect for this role.

\paragraph{Data augmentation plays a crucial role.} Again, because we are in the low-data regime, data augmentation has a much larger impact than architectural changes. Even with non-square maps and complex data masks, adding simple geometric augmentations increases the training loss but substantially reduces and stabilizes the validation loss. We find that considering all symmetries of the dihedral group $D_4$ is also important, in agreement with \citet{Ribli2019}. Missing a subset of these symmetries leads to a noticeable drop in performance, indicating the substantial benefit of any data augmentation for this challenge. As with \citet{Ribli2019}, significant improvements in score are also obtained through test-time augmentation, where the input maps are augmented according to all $D_4$ symmetries and the outputs are averaged for the final prediction.

\subsection{Methodological Insights} \label{sec:methodology}

\paragraph{Agentic systems can accelerate scientific discovery.} Our results show using LLMs and agentic systems can act as a force multiplier within scientific research, enabling a far broader exploration of the method space than by human researchers alone. Within the time constraints of the competition, we were able to match and exceed a team of domain experts in a cosmology inverse problem. Agents allow for cross-domain transfer of ideas, exemplified by the workflow identifying Inception-style CNN architectures for multi-scale feature extraction, posterior inference over the discrete, known cosmology points, and multiple regularization and stability modifications to mean and covariance estimates for the likelihood estimation.

\paragraph{Ensembling LLMs achieves the best results.} We obtained the best results by ensembling the proposals from a diverse set of agentic systems and LLMs, including \texttt{Cmbagent}, \texttt{GPT-5} \citep{singh2025openaigpt5card}, \texttt{Gemini-2.5} \citep{comanici2025gemini25pushingfrontier}, \texttt{Claude-4.5} \citep{claude2026}, and \texttt{Qwen3-max} \citep{yang2025qwen3technicalreport}. We found that overly relying on a single prompt or model results in suggested pipelines clustered around a narrow family of ideas, and querying several LLMs broadened the design space considerably.

\subsection{Limits of autonomous agents}


\paragraph{Agentic systems did not fully replace human experts.} As shown in Section \ref{sec:failure_modes}, the autonomous agentic system did not achieve expert-level performance without human intervention. We believe that this can be rectified or improved in future implementations of \texttt{Cmbagent} with iterative planning, in which $\Phi_\text{plan}$ is updated after each code execution to allow reflections on the results. Furthermore, even if agentic systems advance to the state that they can act fully autonomous, it is necessary for experts to understand and qualitatively assess the outputs of LLMs for these to be trusted by the wider scientific community. We acknowledge that agentic systems raise questions about the nature of doing research and we refer to \citet{hogg2026astrophysics} and \citet{Trotta_2025} for further discussions on this topic.

\paragraph{Optimization needs objective metrics to optimize.} As defined in Section \ref{sec:prob_eval}, we found a pipeline that maximizes some objective scalar metric with which performance can be assessed. Scientific questions often do not have access to this and, more importantly, cannot always be reduced to an optimization problem. More work is required to assess the applicability to a system like this in such settings.

\subsection{Future Research Directions}

While our approach achieved first-place performance in the FAIR Universe Weak Lensing Uncertainty Challenge, moving forward we will focus on further development of the \texttt{Cmbagent} system to aid researchers in accelerating scientific discovery. We aim to do this through adaptive planning with feedback and results from code execution. To increase the flexibility of \texttt{Cmbagent}, we will incorporate the ability to generate specialized sub-agents that are task specific with their own relevant context for the overall problem. For work in this direction, see \texttt{ContextMaker}\footnote{\url{https://github.com/CMBAgents/contextmaker}}. Both these directions will allow \texttt{Cmbagent} to self-improve and evolve its plan and sub-modules during task completion \citep{dang2025multiagentcollaborationevolvingorchestration, gao2026surveyselfevolvingagentswhat}. We also plan to use and demonstrate this system on a wider range of scientific problems and regimes, beyond parameter inference. 

\section{Conclusion}
\label{sec:conclusion}

In this work, we present a case study of agent-driven scientific discovery applied to cosmological parameter inference from weak gravitational lensing. Our central contribution is demonstrating how a multi-agent workflow, combined with lightweight human oversight, can be used to navigate a complex space of modeling and inference choices to assemble an effective pipeline.  This approach led to an inference pipeline that integrates a CNN architecture capable of extracting multi-scale information, symmetry-aware data augmentation, and likelihood calibration, to produce accurate parameter estimates and calibrated uncertainties. 

At the same time, our results highlight clear limitations of fully autonomous agent-based discovery. Without interventions, agentic systems concentrated on performance-saturated ResNet and overlooked better performing alternatives. Incorporating human judgment into the loop was critical to find effective and physically grounded solutions.

While the application here is weak gravitational lensing, the principles and approaches are more general. Agent-driven workflows offer a practical mechanism for structuring and accelerating scientific discovery. When used as tools to expand and test hypotheses in conjunction with expert understanding, such a system can aid scientists to become more efficient and effective.

\section*{Acknowledgments}
We are indebted to all the participants and organizers of the FAIR Universe Challenge, in particular the other winners, including François Lanusse and his team, and Shubhojit Naskar. We thank I\~nigo Zubeldia, Francisco Villaescusa-Navarro, Miles Cranmer and James Fergusson for discussion. Our work is supported by the Infosys-Cambridge AI Centre,  ai@cam and Google.

\section*{Author Contribution}
TB, LX, and AN contributed equally in co-leading the work and preparing the manuscript. TB, LX, AN, BB, SP, EA and CL were participants in the FAIR Universe Challenge on Weak Lensing Uncertainty. BD, PWC and WB prepared the data, coordinated the challenge rules and final leaderboards. Forthcoming publications led by the FAIR Universe organizing team will describe the challenge in detail.

\bibliographystyle{plainnat}
\bibliography{refs}

@ARTICLE{hsc_ssp,
       author = {{Aihara}, Hiroaki and {Arimoto}, Nobuo and {Armstrong}, Robert and {Arnouts}, St{\'e}phane and {Bahcall}, Neta A. and {Bickerton}, Steven and {Bosch}, James and {Bundy}, Kevin and {Capak}, Peter L. and {Chan}, James H.~H. and {Chiba}, Masashi and {Coupon}, Jean and {Egami}, Eiichi and {Enoki}, Motohiro and {Finet}, Francois and {Fujimori}, Hiroki and {Fujimoto}, Seiji and {Furusawa}, Hisanori and {Furusawa}, Junko and {Goto}, Tomotsugu and {Goulding}, Andy and {Greco}, Johnny P. and {Greene}, Jenny E. and {Gunn}, James E. and {Hamana}, Takashi and {Harikane}, Yuichi and {Hashimoto}, Yasuhiro and {Hattori}, Takashi and {Hayashi}, Masao and {Hayashi}, Yusuke and {He{\l}miniak}, Krzysztof G. and {Higuchi}, Ryo and {Hikage}, Chiaki and {Ho}, Paul T.~P. and {Hsieh}, Bau-Ching and {Huang}, Kuiyun and {Huang}, Song and {Ikeda}, Hiroyuki and {Imanishi}, Masatoshi and {Inoue}, Akio K. and {Iwasawa}, Kazushi and {Iwata}, Ikuru and {Jaelani}, Anton T. and {Jian}, Hung-Yu and {Kamata}, Yukiko and {Karoji}, Hiroshi and {Kashikawa}, Nobunari and {Katayama}, Nobuhiko and {Kawanomoto}, Satoshi and {Kayo}, Issha and {Koda}, Jin and {Koike}, Michitaro and {Kojima}, Takashi and {Komiyama}, Yutaka and {Konno}, Akira and {Koshida}, Shintaro and {Koyama}, Yusei and {Kusakabe}, Haruka and {Leauthaud}, Alexie and {Lee}, Chien-Hsiu and {Lin}, Lihwai and {Lin}, Yen-Ting and {Lupton}, Robert H. and {Mandelbaum}, Rachel and {Matsuoka}, Yoshiki and {Medezinski}, Elinor and {Mineo}, Sogo and {Miyama}, Shoken and {Miyatake}, Hironao and {Miyazaki}, Satoshi and {Momose}, Rieko and {More}, Anupreeta and {More}, Surhud and {Moritani}, Yuki and {Moriya}, Takashi J. and {Morokuma}, Tomoki and {Mukae}, Shiro and {Murata}, Ryoma and {Murayama}, Hitoshi and {Nagao}, Tohru and {Nakata}, Fumiaki and {Niida}, Mana and {Niikura}, Hiroko and {Nishizawa}, Atsushi J. and {Obuchi}, Yoshiyuki and {Oguri}, Masamune and {Oishi}, Yukie and {Okabe}, Nobuhiro and {Okamoto}, Sakurako and {Okura}, Yuki and {Ono}, Yoshiaki and {Onodera}, Masato and {Onoue}, Masafusa and {Osato}, Ken and {Ouchi}, Masami and {Price}, Paul A. and {Pyo}, Tae-Soo and {Sako}, Masao and {Sawicki}, Marcin and {Shibuya}, Takatoshi and {Shimasaku}, Kazuhiro and {Shimono}, Atsushi and {Shirasaki}, Masato and {Silverman}, John D. and {Simet}, Melanie and {Speagle}, Joshua and {Spergel}, David N. and {Strauss}, Michael A. and {Sugahara}, Yuma and {Sugiyama}, Naoshi and {Suto}, Yasushi and {Suyu}, Sherry H. and {Suzuki}, Nao and {Tait}, Philip J. and {Takada}, Masahiro and {Takata}, Tadafumi and {Tamura}, Naoyuki and {Tanaka}, Manobu M. and {Tanaka}, Masaomi and {Tanaka}, Masayuki and {Tanaka}, Yoko and {Terai}, Tsuyoshi and {Terashima}, Yuichi and {Toba}, Yoshiki and {Tominaga}, Nozomu and {Toshikawa}, Jun and {Turner}, Edwin L. and {Uchida}, Tomohisa and {Uchiyama}, Hisakazu and {Umetsu}, Keiichi and {Uraguchi}, Fumihiro and {Urata}, Yuji and {Usuda}, Tomonori and {Utsumi}, Yousuke and {Wang}, Shiang-Yu and {Wang}, Wei-Hao and {Wong}, Kenneth C. and {Yabe}, Kiyoto and {Yamada}, Yoshihiko and {Yamanoi}, Hitomi and {Yasuda}, Naoki and {Yeh}, Sherry and {Yonehara}, Atsunori and {Yuma}, Suraphong},
        title = "{The Hyper Suprime-Cam SSP Survey: Overview and survey design}",
      journal = {Publ. Astron. Soc. Jpn.},
         year = 2018,
        month = jan,
       volume = {70},
          eid = {S4},
        pages = {S4},
          doi = {10.1093/pasj/psx066},
archivePrefix = {arXiv},
       eprint = {1704.05858},
 primaryClass = {astro-ph.IM},
       adsurl = {https://ui.adsabs.harvard.edu/abs/2018PASJ...70S...4A}
}

@ARTICLE{hsc_y3,
       author = {{More}, Surhud and {Sugiyama}, Sunao and {Miyatake}, Hironao and {Rau}, Markus Michael and {Shirasaki}, Masato and {Li}, Xiangchong and {Nishizawa}, Atsushi J. and {Osato}, Ken and {Zhang}, Tianqing and {Takada}, Masahiro and {Hamana}, Takashi and {Takahashi}, Ryuichi and {Dalal}, Roohi and {Mandelbaum}, Rachel and {Strauss}, Michael A. and {Kobayashi}, Yosuke and {Nishimichi}, Takahiro and {Oguri}, Masamune and {Luo}, Wentao and {Kannawadi}, Arun and {Hsieh}, Bau-Ching and {Armstrong}, Robert and {Bosch}, James and {Komiyama}, Yutaka and {Lupton}, Robert H. and {Lust}, Nate B. and {MacArthur}, Lauren A. and {Miyazaki}, Satoshi and {Murayama}, Hitoshi and {Okura}, Yuki and {Price}, Paul A. and {Tait}, Philip J. and {Tanaka}, Masayuki and {Wang}, Shiang-Yu},
        title = "{Hyper Suprime-Cam Year 3 results: Measurements of clustering of SDSS-BOSS galaxies, galaxy-galaxy lensing, and cosmic shear}",
      journal = {Phys. Rev. D},
         year = 2023,
        month = dec,
       volume = {108},
       number = {12},
          eid = {123520},
        pages = {123520},
          doi = {10.1103/PhysRevD.108.123520},
archivePrefix = {arXiv},
       eprint = {2304.00703},
 primaryClass = {astro-ph.CO},
       adsurl = {https://ui.adsabs.harvard.edu/abs/2023PhRvD.108l3520M}
}

@misc{casas2025clappclassllmagent,
      title={CLAPP: The CLASS LLM Agent for Pair Programming}, 
      author={Santiago Casas and Christian Fidler and Boris Bolliet and Francisco Villaescusa-Navarro and Julien Lesgourgues},
      year={2025},
      eprint={2508.05728},
      archivePrefix={arXiv},
      primaryClass={astro-ph.IM},
      url={https://arxiv.org/abs/2508.05728}, 
}

@article{Kilbinger_2015,
   title={Cosmology with cosmic shear observations: a review},
   volume={78},
   ISSN={1361-6633},
   url={http://dx.doi.org/10.1088/0034-4885/78/8/086901},
   DOI={10.1088/0034-4885/78/8/086901},
   number={8},
   journal={Reports on Progress in Physics},
   publisher={IOP Publishing},
   author={Kilbinger, Martin},
   year={2015},
   month=jul, pages={086901}, }

@misc{comanici2025gemini25pushingfrontier,
      title={Gemini 2.5: Pushing the Frontier with Advanced Reasoning, Multimodality, Long Context, and Next Generation Agentic Capabilities}, 
      author={Gheorghe Comanici and Eric Bieber and Mike Schaekermann and Ice Pasupat and others},
      year={2025},
      eprint={2507.06261},
      archivePrefix={arXiv},
      primaryClass={cs.CL},
      url={https://arxiv.org/abs/2507.06261}, 
}

@misc{singh2025openaigpt5card,
      title={OpenAI GPT-5 System Card}, 
      author={Aaditya Singh and Adam Fry and Adam Perelman and Adam Tart and Adi Ganesh and Ahmed El-Kishky and others},
      year={2025},
      eprint={2601.03267},
      archivePrefix={arXiv},
      primaryClass={cs.CL},
      url={https://arxiv.org/abs/2601.03267}, 
}

@misc{yang2025qwen3technicalreport,
      title={Qwen3 Technical Report}, 
      author={An Yang and Anfeng Li and Baosong Yang and Beichen Zhang and Binyuan Hui and Bo Zheng and Bowen Yu and Chang Gao and Chengen Huang and Chenxu Lv and Chujie Zheng and Dayiheng Liu and Fan Zhou and others},
      year={2025},
      eprint={2505.09388},
      archivePrefix={arXiv},
      primaryClass={cs.CL},
      url={https://arxiv.org/abs/2505.09388}, 
}

@inproceedings{mudur2025llm,
  title={An LLM-driven framework for cosmological model-building and exploration},
  author={Mudur, Nayantara and Cuesta-Lazaro, Carolina and Toomey, Michael W. and Finkbeiner, Douglas P.},
  booktitle={COLM 2025 Workshop on Language Models for Science (LM4Sci)},
  year={2025},
  url={https://openreview.net/forum?id=xnPJOtPmK3}
}

@article{Ribli2019,
    author = {Ribli, Dezső and Pataki, Bálint Ármin and Zorrilla Matilla, José Manuel and Hsu, Daniel and Haiman, Zoltán and Csabai, István},
    title = {Weak lensing cosmology with convolutional neural networks on noisy data},
    journal = {Monthly Notices of the Royal Astronomical Society},
    volume = {490},
    number = {2},
    pages = {1843-1860},
    year = {2019},
    month = {09},
    issn = {0035-8711},
    doi = {10.1093/mnras/stz2610},
    url = {https://doi.org/10.1093/mnras/stz2610},
    eprint = {https://academic.oup.com/mnras/article-pdf/490/2/1843/30194757/stz2610.pdf},
}

@ARTICLE{LSST2026,
       author = {{LSST Dark Energy Science Collaboration} and {Aubourg}, Eric and {Avestruz}, Camille and {Becker}, Matthew R. and {Biswas}, Biswajit and {Biswas}, Rahul and {Bolliet}, Boris and {Bolton}, Adam S. and {Bom}, Clecio R. and {Bonnet-Guerrini}, Rapha{\"e}l and {Boucaud}, Alexandre and {Campagne}, Jean-Eric and {Chang}, Chihway and {{\'C}iprijanovi{\'c}}, Aleksandra and {Cohen-Tanugi}, Johann and {Coughlin}, Michael W. and {Crenshaw}, John Franklin and {Cuevas-Tello}, Juan C. and {de Vicente}, Juan and {Digel}, Seth W. and {Dillmann}, Steven and {de Le{\'o}n Dominguez Romero}, Mariano Javier and {Drlica-Wagner}, Alex and {Erickson}, Sydney and {Gagliano}, Alexander T. and {Georgiou}, Christos and {Ghosh}, Aritra and {Grayling}, Matthew and {Grishin}, Kirill A. and {Heavens}, Alan and {House}, Lindsay R. and {Ishak}, Mustapha and {Kabalan}, Wassim and {Kannawadi}, Arun and {Lanusse}, Fran{\c{c}}ois and {Leonard}, C. Danielle and {L{\'e}get}, Pierre-Fran{\c{c}}ois and {Lochner}, Michelle and {Mao}, Yao-Yuan and {Melchior}, Peter and {Merz}, Grant and {Millon}, Martin and {M{\"o}ller}, Anais and {Narayan}, Gautham and {Omori}, Yuuki and {Peiris}, Hiranya and {Perreault-Levasseur}, Laurence and {Plazas Malag{\'o}n}, Andr{\'e}s A. and {Ramachandra}, Nesar and {Remy}, Benjamin and {Roucelle}, C{\'e}cile and {Ruiz-Zapatero}, Jaime and {Schuldt}, Stefan and {Sevilla-Noarbe}, Ignacio and {Shah}, Ved G. and {Starkenburg}, Tjitske and {Thorp}, Stephen and {Toribio San Cipriano}, Laura and {Tr{\"o}ster}, Tilman and {Trotta}, Roberto and {Venkatraman}, Padma and {Wasserman}, Amanda and {White}, Tim and {Zeghal}, Justine and {Zhang}, Tianqing and {Zhang}, Yuanyuan},
        title = "{Opportunities in AI/ML for the Rubin LSST Dark Energy Science Collaboration}",
      journal = {arXiv e-prints},
         year = 2026,
        month = jan,
          eid = {arXiv:2601.14235},
        pages = {arXiv:2601.14235},
          doi = {10.48550/arXiv.2601.14235},
archivePrefix = {arXiv},
       eprint = {2601.14235},
 primaryClass = {astro-ph.IM},
       adsurl = {https://ui.adsabs.harvard.edu/abs/2026arXiv260114235L}
}

@INPROCEEDINGS{he2016,
       author = {{He}, Kaiming and {Zhang}, Xiangyu and {Ren}, Shaoqing and {Sun}, Jian},
        title = "{Deep Residual Learning for Image Recognition}",
    booktitle = {2016 IEEE Conference on Computer Vision and Pattern Recognition (CVPR)},
         year = 2016,
        month = jun,
          eid = {1},
        pages = {1},
          doi = {10.1109/CVPR.2016.90},
archivePrefix = {arXiv},
       eprint = {1512.03385},
 primaryClass = {cs.CV},
       adsurl = {https://ui.adsabs.harvard.edu/abs/2016cvpr.confE...1H}
}

@misc{gao2026surveyselfevolvingagentswhat,
      title={A Survey of Self-Evolving Agents: What, When, How, and Where to Evolve on the Path to Artificial Super Intelligence}, 
      author={{Huan-ang} Gao and Jiayi Geng and Wenyue Hua and Mengkang Hu and Xinzhe Juan and Hongzhang Liu and Shilong Liu and Jiahao Qiu and Xuan Qi and Yiran Wu and Hongru Wang and Han Xiao and Yuhang Zhou and Shaokun Zhang and Jiayi Zhang and Jinyu Xiang and Yixiong Fang and Qiwen Zhao and Dongrui Liu and Qihan Ren and Cheng Qian and Zhenhailong Wang and Minda Hu and Huazheng Wang and Qingyun Wu and Heng Ji and Mengdi Wang},
      year={2026},
      eprint={2507.21046},
      archivePrefix={arXiv},
      primaryClass={cs.AI},
      url={https://arxiv.org/abs/2507.21046}, 
}

@article{Ribli_2019,
   title={Weak lensing cosmology with convolutional neural networks on noisy data},
   volume={490},
   ISSN={1365-2966},
   url={http://dx.doi.org/10.1093/mnras/stz2610},
   DOI={10.1093/mnras/stz2610},
   number={2},
   journal={Monthly Notices of the Royal Astronomical Society},
   publisher={Oxford University Press (OUP)},
   author={Ribli, Dezső and Pataki, Bálint Ármin and Zorrilla Matilla, José Manuel and Hsu, Daniel and Haiman, Zoltán and Csabai, István},
   year={2019},
   month=sep, pages={1843–1860} }

@article{PhysRevD.110.043535,
  title = {Improving convolutional neural networks for cosmological fields with random permutation},
  author = {Zhong, Kunhao and Gatti, Marco and Jain, Bhuvnesh},
  journal = {Phys. Rev. D},
  volume = {110},
  issue = {4},
  pages = {043535},
  numpages = {18},
  year = {2024},
  month = {Aug},
  publisher = {American Physical Society},
  doi = {10.1103/PhysRevD.110.043535},
  url = {https://link.aps.org/doi/10.1103/PhysRevD.110.043535}
}

@misc{dang2025multiagentcollaborationevolvingorchestration,
      title={Multi-Agent Collaboration via Evolving Orchestration}, 
      author={Yufan Dang and Chen Qian and Xueheng Luo and Jingru Fan and Zihao Xie and Ruijie Shi and Weize Chen and Cheng Yang and Xiaoyin Che and Ye Tian and Xuantang Xiong and Lei Han and Zhiyuan Liu and Maosong Sun},
      year={2025},
      eprint={2505.19591},
      archivePrefix={arXiv},
      primaryClass={cs.CL},
      url={https://arxiv.org/abs/2505.19591}, 
}

@misc{moss2025aicosmologistiagentic,
      title={The AI Cosmologist I: An Agentic System for Automated Data Analysis}, 
      author={Adam Moss},
      year={2025},
      eprint={2504.03424},
      archivePrefix={arXiv},
      primaryClass={astro-ph.IM},
      url={https://arxiv.org/abs/2504.03424}, 
}

@ARTICLE{wei2025,
       author = {{Wei}, Jiaqi and {Yang}, Yuejin and {Zhang}, Xiang and {Chen}, Yuhan and {Zhuang}, Xiang and {Gao}, Zhangyang and {Zhou}, Dongzhan and {Wang}, Guangshuai and {Gao}, Zhiqiang and {Cao}, Juntai and {Qiu}, Zijie and {Hu}, Ming and {Ma}, Chenglong and {Tang}, Shixiang and {He}, Junjun and {Song}, Chunfeng and {He}, Xuming and {Zhang}, Qiang and {You}, Chenyu and {Zheng}, Shuangjia and {Ding}, Ning and {Ouyang}, Wanli and {Dong}, Nanqing and {Cheng}, Yu and {Sun}, Siqi and {Bai}, Lei and {Zhou}, Bowen},
        title = "{From AI for Science to Agentic Science: A Survey on Autonomous Scientific Discovery}",
      journal = {arXiv e-prints},
         year = 2025,
        month = aug,
          eid = {arXiv:2508.14111},
        pages = {arXiv:2508.14111},
          doi = {10.48550/arXiv.2508.14111},
archivePrefix = {arXiv},
       eprint = {2508.14111},
 primaryClass = {stat.ML},
       adsurl = {https://ui.adsabs.harvard.edu/abs/2025arXiv250814111W}
}

@software{CMBAGENT_2025,
            author = {Boris Bolliet},
            title = {CMBAGENT: Open-Source Multi-Agent System for Science},
            year = {2025},
            url = {https://github.com/CMBAgents/cmbagent},
            note = {Available at https://github.com/CMBAgents/cmbagent},
            version = {latest}
            }

@misc{xu2025opensourceplanning,
    title={Open Source Planning \& Control System with Language Agents for Autonomous Scientific Discovery},
    author={Licong Xu and Milind Sarkar and Anto I. Lonappan and Íñigo Zubeldia and Pablo Villanueva-Domingo and Santiago Casas and Christian Fidler and Chetana Amancharla and Ujjwal Tiwari and Adrian Bayer and Chadi Ait Ekiou and Miles Cranmer and Adrian Dimitrov and James Fergusson and Kahaan Gandhi and Sven Krippendorf and Andrew Laverick and Julien Lesgourgues and Antony Lewis and Thomas Meier and Blake Sherwin and Kristen Surrao and Francisco Villaescusa-Navarro and Chi Wang and Xueqing Xu and Boris Bolliet},
    year={2025},
    eprint={2507.07257},
    archivePrefix={arXiv},
    primaryClass={cs.AI},
    url={https://arxiv.org/abs/2507.07257},
}

@misc{Laverick:2024fyh,
  author = "Laverick, Andrew and Surrao, Kristen and Zubeldia, Inigo and Bolliet, Boris and Cranmer, Miles and Lewis, Antony and Sherwin, Blake and Lesgourgues, Julien",
  title = "{Multi-Agent System for Cosmological Parameter Analysis}",
  eprint = "2412.00431",
  archivePrefix = "arXiv",
  primaryClass = "astro-ph.IM",
  month = "11",
  year = "2024"
}

@misc{claude2026,
  author = {{Anthropic}},
  title = {Claude 4.5 Opus},
  year = {2026},
  note = {Large language model},
  url = {https://www.anthropic.com},
  urldate = {2026-03-06}
}

@misc{ting2025egentautonomousagentequivalent,
      title={Egent: An Autonomous Agent for Equivalent Width Measurement}, 
      author={Yuan-Sen Ting and Serat Mahmud Saad and Fan Liu and Yuting Shen},
      year={2025},
      eprint={2512.01270},
      archivePrefix={arXiv},
      primaryClass={astro-ph.IM},
      url={https://arxiv.org/abs/2512.01270}, 
}

@misc{pinheiro2025largelanguagemodelsachieve,
      title={Large Language Models Achieve Gold Medal Performance at the International Olympiad on Astronomy \& Astrophysics (IOAA)}, 
      author={Lucas Carrit Delgado Pinheiro and Ziru Chen and Bruno Caixeta Piazza and Ness Shroff and Yingbin Liang and Yuan-Sen Ting and Huan Sun},
      year={2025},
      eprint={2510.05016},
      archivePrefix={arXiv},
      primaryClass={astro-ph.IM},
      url={https://arxiv.org/abs/2510.05016}, 
}

@misc{ramachandra2025teachingllmsspeakspectroscopy,
      title={Teaching LLMs to Speak Spectroscopy}, 
      author={Nesar Ramachandra and Yuan-Sen Ting and Zechang Sun and Azton Wells and Salman Habib},
      year={2025},
      eprint={2508.10075},
      archivePrefix={arXiv},
      primaryClass={astro-ph.IM},
      url={https://arxiv.org/abs/2508.10075}, 
}

@misc{li2026madevolveevolutionaryoptimizationcosmological,
      title={MadEvolve: Evolutionary Optimization of Cosmological Algorithms with Large Language Models}, 
      author={Tianyi Li and Shihui Zang and Moritz Münchmeyer},
      year={2026},
      eprint={2602.15951},
      archivePrefix={arXiv},
      primaryClass={astro-ph.CO},
      url={https://arxiv.org/abs/2602.15951}, 
}

@misc{gao2025testtimescalingtechniquestheoretical,
      title={Test-time Scaling Techniques in Theoretical Physics -- A Comparison of Methods on the TPBench Dataset}, 
      author={Zhiqi Gao and Tianyi Li and Yurii Kvasiuk and Sai Chaitanya Tadepalli and Maja Rudolph and Daniel J. H. Chung and Frederic Sala and Moritz Münchmeyer},
      year={2025},
      eprint={2506.20729},
      archivePrefix={arXiv},
      primaryClass={cs.LG},
      url={https://arxiv.org/abs/2506.20729}, 
}

@misc{ghafarollahi2024sciagentsautomatingscientificdiscovery,
      title={SciAgents: Automating scientific discovery through multi-agent intelligent graph reasoning}, 
      author={Alireza Ghafarollahi and Markus J. Buehler},
      year={2024},
      eprint={2409.05556},
      archivePrefix={arXiv},
      primaryClass={cs.AI},
      url={https://arxiv.org/abs/2409.05556}, 
}

@misc{sun2025mephistoselfimprovinglargelanguage,
      title={Mephisto: Self-Improving Large Language Model-Based Agents for Automated Interpretation of Multi-band Galaxy Observations}, 
      author={Zechang Sun and Yuan-Sen Ting and Yaobo Liang and Nan Duan and Song Huang and Zheng Cai},
      year={2025},
      eprint={2510.08354},
      archivePrefix={arXiv},
      primaryClass={astro-ph.IM},
      url={https://arxiv.org/abs/2510.08354}, 
}

@misc{peng2026deepinflationaiagentresearch,
      title={DeepInflation: an AI agent for research and model discovery of inflation}, 
      author={Ze-Yu Peng and Hao-Shi Yuan and Qi Lai and Jun-Qian Jiang and Gen Ye and Jun Zhang and Yun-Song Piao},
      year={2026},
      eprint={2601.14288},
      archivePrefix={arXiv},
      primaryClass={astro-ph.CO},
      url={https://arxiv.org/abs/2601.14288}, 
}

@misc{hogg2026astrophysics,
      title={Why do we do astrophysics?}, 
      author={David W. Hogg},
      year={2026},
      eprint={2602.10181},
      archivePrefix={arXiv},
      primaryClass={astro-ph.IM},
      url={https://arxiv.org/abs/2602.10181}, 
}

@article{Trotta_2025,
   title={The indiscriminate adoption of AI threatens the foundations of academia},
   volume={9},
   ISSN={2397-3366},
   url={http://dx.doi.org/10.1038/s41550-025-02738-w},
   DOI={10.1038/s41550-025-02738-w},
   number={12},
   journal={Nature Astronomy},
   publisher={Springer Science and Business Media LLC},
   author={Trotta, Roberto},
   year={2025},
   month=dec, pages={1748–1749} }

@misc{miao2025physmasterbuildingautonomousai,
      title={PhysMaster: Building an Autonomous AI Physicist for Theoretical and Computational Physics Research}, 
      author={Tingjia Miao and Jiawen Dai and Jingkun Liu and Jinxin Tan and Muhua Zhang and Wenkai Jin and Yuwen Du and Tian Jin and Xianghe Pang and Zexi Liu and Tu Guo and Zhengliang Zhang and Yunjie Huang and Shuo Chen and Rui Ye and Yuzhi Zhang and Linfeng Zhang and Kun Chen and Wei Wang and Weinan E and Siheng Chen},
      year={2025},
      eprint={2512.19799},
      archivePrefix={arXiv},
      primaryClass={cs.AI},
      url={https://arxiv.org/abs/2512.19799}, 
}

@misc{arlt2025autonomousquantumphysicsresearch,
      title={Towards autonomous quantum physics research using LLM agents with access to intelligent tools}, 
      author={Sören Arlt and Xuemei Gu and Mario Krenn},
      year={2025},
      eprint={2511.11752},
      archivePrefix={arXiv},
      primaryClass={cs.AI},
      url={https://arxiv.org/abs/2511.11752}, 
}

@misc{zhang2025bridgingliteratureuniversemultiagent,
      title={Bridging Literature and the Universe Via A Multi-Agent Large Language Model System}, 
      author={Xiaowen Zhang and Zhenyu Bi and Patrick Lachance and Xuan Wang and Tiziana Di Matteo and Rupert A. C. Croft},
      year={2025},
      eprint={2507.08958},
      archivePrefix={arXiv},
      primaryClass={astro-ph.IM},
      url={https://arxiv.org/abs/2507.08958}, 
}

@misc{ye2025replicationbenchaiagentsreplicate,
      title={ReplicationBench: Can AI Agents Replicate Astrophysics Research Papers?}, 
      author={Christine Ye and Sihan Yuan and Suchetha Cooray and Steven Dillmann and Ian L. V. Roque and Dalya Baron and Philipp Frank and Sergio Martin-Alvarez and Nolan Koblischke and Frank J Qu and Diyi Yang and Risa Wechsler and Ioana Ciuca},
      year={2025},
      eprint={2510.24591},
      archivePrefix={arXiv},
      primaryClass={cs.CL},
      url={https://arxiv.org/abs/2510.24591}, 
}

@misc{nagele2026agenticexplorationphysicsmodels,
      title={Agentic Exploration of Physics Models}, 
      author={Maximilian Nägele and Florian Marquardt},
      year={2026},
      eprint={2509.24978},
      archivePrefix={arXiv},
      primaryClass={cs.AI},
      url={https://arxiv.org/abs/2509.24978}, 
}

@article{denario,
         title={The Denario project: Deep knowledge AI agents for scientific discovery}, 
         author={Francisco Villaescusa-Navarro and Boris Bolliet and Pablo Villanueva-Domingo and Adrian E. Bayer and Aidan Acquah and Chetana Amancharla and Almog Barzilay-Siegal and Pablo Bermejo and Camille Bilodeau and Pablo Cárdenas Ramírez and Miles Cranmer and Urbano L. França and ChangHoon Hahn and Yan-Fei Jiang and Raul Jimenez and Jun-Young Lee and Antonio Lerario and Osman Mamun and Thomas Meier and Anupam A. Ojha and Pavlos Protopapas and Shimanto Roy and David N. Spergel and Pedro Tarancón-Álvarez and Ujjwal Tiwari and Matteo Viel and Digvijay Wadekar and Chi Wang and Bonny Y. Wang and Licong Xu and Yossi Yovel and Shuwen Yue and Wen-Han Zhou and Qiyao Zhu and Jiajun Zou and Íñigo Zubeldia},
         year={2025},
         eprint={2510.26887},
         journal={arXiv e-prints},
         archivePrefix={arXiv},
         primaryClass={cs.AI},
         url={https://arxiv.org/abs/2510.26887},
}

@misc{szegedy2014goingdeeperconvolutions,
      title={Going Deeper with Convolutions}, 
      author={Christian Szegedy and Wei Liu and Yangqing Jia and Pierre Sermanet and Scott Reed and Dragomir Anguelov and Dumitru Erhan and Vincent Vanhoucke and Andrew Rabinovich},
      year={2014},
      eprint={1409.4842},
      archivePrefix={arXiv},
      primaryClass={cs.CV},
      url={https://arxiv.org/abs/1409.4842}, 
}

@article{Hartlap_2006,
   title={Why your model parameter confidences might be too optimistic.  Unbiased estimation of the inverse covariance matrix},
   volume={464},
   ISSN={1432-0746},
   url={http://dx.doi.org/10.1051/0004-6361:20066170},
   DOI={10.1051/0004-6361:20066170},
   number={1},
   journal={Astronomy \& Astrophysics},
   publisher={EDP Sciences},
   author={Hartlap, J. and Simon, P. and Schneider, P.},
   year={2006},
   month=dec, pages={399–404} }

@article{LedoitWolf2003,
  author       = {Ledoit, Olivier and Wolf, Michael},
  title        = {Improved estimation of the covariance matrix of stock returns with an application to portfolio selection},
  journal      = {Journal of Empirical Finance},
  year         = {2003},
  volume       = {10},
  number       = {5},
  pages        = {603--621},
  doi          = {10.1016/S0927-5398(03)00007-0},
  url          = {https://doi.org/10.1016/S0927-5398(03)00007-0}
}

@article{Cheng2024,
    author = {Cheng, Sihao and Morel, Rudy and Allys, Erwan and Ménard, Brice and Mallat, Stéphane},
    title = {Scattering spectra models for physics},
    journal = {PNAS Nexus},
    volume = {3},
    number = {4},
    pages = {pgae103},
    year = {2024},
    month = {03},
    issn = {2752-6542},
    doi = {10.1093/pnasnexus/pgae103},
    url = {https://doi.org/10.1093/pnasnexus/pgae103},
    eprint = {https://academic.oup.com/pnasnexus/article-pdf/3/4/pgae103/57336291/pgae103.pdf},
}

\appendix

\section{Detailed Inference Protocol and Hyperparameters} \label{app:inference-details}

\begin{tcolorbox}[title=Algorithm 1: Final inference pipeline,colback=white,colframe=black]\label{algo:1}
\begin{enumerate}
    \item Train an ensemble of CNN models, each on a different train/validation split.
    \item For each ensemble member $m$, compute validation prediction--truth pairs
    $
    \{(\hat{\theta}^{(m)}_i,\theta_i)\}_{i \in I_{\mathrm{val}}^{(m)}}.
    $
    \item For each cosmology grid point $\theta_g$, group the validation predictions with ground truth $\theta_i=\theta_g$ and estimate the mean and covariance,
    $
    \mu_g,\; \Sigma_g.
    $
    \item Define the empirical Gaussian likelihood
    $
    p(\hat{\theta}\mid \theta_g)\approx \mathcal{N}(\hat{\theta};\mu_g,\Sigma_g),
    $
    using a Hartlap-corrected inverse covariance.
    \item Apply $D_4$ test-time augmentation and average the resulting predictions before likelihood evaluation.
    \item Smooth $(\mu_g,\Sigma_g)$ across nearby grid points and regularize the covariance estimates via shrinkage to obtain calibrated moments.
    \item Determine a global temperature $\tau$ from the validation residuals and rescale the covariance matrices accordingly.
    \item Compute unsupervised NLL-based weights for the ensemble members and form the weighted prediction
    $
    \hat{\theta}_{\mathrm{ens}}=\sum_m w_m^{(\mathrm{ens})}\hat{\theta}^{(m)}.
    $
    \item Evaluate the calibrated likelihood over all grid points,
    $
    \tilde{w}_g \propto p(\hat{\theta}_{\mathrm{ens}} \mid \theta_g),
    $
    normalize to obtain $w_g$, and compute
    $
    \hat{\theta}_{\mathrm{post}}=\sum_g w_g\theta_g
    $
    together with the marginal posterior uncertainties.
\end{enumerate}
\end{tcolorbox}

In this appendix, we give the full details of our inference. Our agentic workflow suggested this as one of many alternative approaches to an MCMC pipeline, and we chose this method because it delivered substantially better accuracy and calibration.

\subsection{The Pipeline}

For each ensemble member $m$, we run the trained model on the same held-out validation maps used during training (which are different for each member), generating a collection of prediction--truth pairs
\[
\left\{\, \hat{\boldsymbol{\theta}}_i^{(m)},\ \boldsymbol{\theta}_i \,\right\}_{i \in \mathcal{I}^{(m)}_{\rm val}},
\]
where $\hat{\boldsymbol{\theta}}_i^{(m)} \in \mathbb{R}^2$ denotes the cosmological parameters predicted by member $m$ for validation sample $i$, and $\boldsymbol{\theta}_i = (\Omega_{\rm m}, S_8)$ is the corresponding ground-truth cosmology of that sample. The index set $\mathcal{I}^{(m)}_{\rm val}$ contains all validation maps assigned to ensemble member $m$ by its internal training split. These predictions are then grouped according to their true cosmology point $\boldsymbol{\theta}_g$. This allows us to empirically characterize, at each point, the predicted mean and covariance, given by:
\[
\boldsymbol{\mu}_g
= \frac{1}{N_g} \sum_{i \in G_g} \hat{\boldsymbol{\theta}}_i,
\]
\[
\mathbf{C}_g
= \frac{1}{N_g - 1}
\sum_{i \in G_g}
\left(\hat{\boldsymbol{\theta}}_i - \boldsymbol{\mu}_g\right)
\left(\hat{\boldsymbol{\theta}}_i - \boldsymbol{\mu}_g\right)^{\!\top},
\]
where $G_g$ denotes the set of validation indices whose ground-truth cosmology
is $\boldsymbol{\theta}_g$ and $N_g$ is the number of samples at that
cosmology. We use this empirical covariance to characterise the distribution of
CNN predictions at fixed cosmology. When constructing the Gaussian likelihood,
we apply a Hartlap correction \citep{Hartlap_2006} to the inverse covariance,
\[
\mathbf{\Sigma}_g^{-1}
= \alpha_{\rm H}\,\mathbf{C}_g^{-1},
\qquad
\alpha_{\rm H} = \frac{N_g - d - 2}{N_g - 1},
\]
with $d=2$ the dimensionality of $\boldsymbol{\theta}$.  
The pair $\big(\boldsymbol{\mu}_g,\, \mathbf{\Sigma}_g\big)$ then defines our
Gaussian likelihood model,
\[
p(\hat{\boldsymbol{\theta}}\mid\boldsymbol{\theta}_g)
\simeq
\mathcal{N}\!\left(
\boldsymbol{\mu}_g,\;
\mathbf{\Sigma}_g
\right).
\]
During both validation and test-time inference, we apply test-time augmentation (using all symmetries of $D_4$ dihedral group) and average the corresponding predictions in parameter space before feeding them into the likelihood calibration.

Although the calibration above yields an empirical mean $\boldsymbol{\mu}_g$ and covariance $\mathbf{\Sigma}_g$ at each cosmology grid point, these estimates can be noisy when $N_g$ is finite, especially where the coverage of our data is low. To enforce smoothness of the likelihood across neighboring cosmologies, we apply a kernel-smoothing procedure in the $(\Omega_{\rm m}, S_8)$ parameter space. For each point $g$, we replace the raw empirical moments $\{\boldsymbol{\mu}_g, \mathbf{\Sigma}_g\}$ with locally averaged quantities, 
\[
\bar{\boldsymbol{\mu}}_g
= \sum_{g'} W_{g g'}\, \boldsymbol{\mu}_{g'},
\qquad
\bar{\mathbf{\Sigma}}_g
= \sum_{g'} W_{g g'}\,
\Big[
\mathbf{\Sigma}_{g'}
+ (\boldsymbol{\mu}_{g'} - \bar{\boldsymbol{\mu}}_g)
  (\boldsymbol{\mu}_{g'} - \bar{\boldsymbol{\mu}}_g)^{\!\top}
\Big],
\]
where $W_{g g'}$ is a Gaussian kernel in parameter space, normalized so that $\sum_{g'} W_{g g'} = 1$. The Gaussian kernel has standard deviation equal to $\sigma_{\rm BW} \cdot \textrm{med}_5$, where $\sigma_{\rm BW}$ is a scaling factor and $\textrm{med}_5$ is the median distance to the fifth nearest neighbor across all points in the grid (i.e.,\, a natural measure of the typical grid spacing). This smoothing aggregates information from nearby cosmology points, suppressing statistical fluctuations in the per-grid estimates while preserving the overall structure of the likelihood. The resulting smoothed moments $(\bar{\boldsymbol{\mu}}_g, \bar{\mathbf{\Sigma}}_g)$ therefore provide a more stable and better-calibrated approximation to the distribution of CNN predictions, especially in regions where the validation set
is sparse.

To further stabilize the covariance estimates after kernel smoothing, we apply a mild Ledoit--Wolf--style shrinkage \citep{LedoitWolf2003} to each smoothed covariance matrix. This reduces spurious off-diagonal correlations that may arise from finite validation
samples while preserving the overall scaling of the uncertainties. We also introduce a global temperature-scaling factor $\tau$, chosen such that the distribution of whitened residuals across all validation predictions matches the expected $\chi^2$ distribution with $p_{\rm DOF}=2$ degrees of freedom. The resulting calibrated covariance is
\[
\widetilde{\mathbf{\Sigma}}_g
= \tau^{2}
\Big[(1-\lambda_{\rm LW})\,\bar{\mathbf{\Sigma}}_g
      + \lambda_{\rm LW}\,\mathrm{diag}(\bar{\mathbf{\Sigma}}_g)\Big],
\]
where $\lambda_{\rm LW}$ is the shrinkage amplitude. These fully calibrated moments $(\bar{\boldsymbol{\mu}}_g,\,\widetilde{\mathbf{\Sigma}}_g)$ define the final Gaussian likelihood model used for inference. The determination of $\tau$ is shown in Appendix \ref{app:tau}. In practice, these calibrated moments $(\bar{\boldsymbol{\mu}}_g,\,\widetilde{\mathbf{\Sigma}}_g)$ are estimated from the \emph{concatenated} validation predictions of all selected ensemble members, so that a single global likelihood model is used for all networks.

We adopt a uniform prior over the cosmology grid, so that the posterior is determined entirely by the normalized likelihood weights. Given an observed CNN prediction $\hat{\boldsymbol{\theta}}_{\rm obs}$, we evaluate the calibrated Gaussian likelihood at each grid point,
\[
\tilde{w}_g \propto
\exp\!\left[
-\tfrac{1}{2}
\bigl(\hat{\boldsymbol{\theta}}_{\rm obs}-\bar{\boldsymbol{\mu}}_g\bigr)^{\!\top}
\widetilde{\mathbf{\Sigma}}_g^{-1}
\bigl(\hat{\boldsymbol{\theta}}_{\rm obs}-\bar{\boldsymbol{\mu}}_g\bigr)
\right],
\]
and obtain the discrete posterior by normalization,
\[
w_g = \frac{\tilde{w}_g}{\sum_{g'} \tilde{w}_{g'}}.
\]
The posterior mean estimator is then
\[
\hat{\boldsymbol{\theta}}_{\rm post}
= \sum_g w_g\,\boldsymbol{\theta}_g,
\]
with marginal uncertainties given by
\[
\sigma^2_{\rm post}(\theta_a)
= \sum_g w_g\,\bigl(\theta_{g,a}-\hat{\theta}_{{\rm post},a}\bigr)^2,
\qquad a\in\{\Omega_{\rm m},S_8\}.
\]
We aggregate the outputs of the 10 independently trained CNNs into an ensemble. For each member, we compute predictions on the test maps and evaluate their consistency with the calibrated likelihood via an unsupervised negative log-likelihood (NLL). Members with lower NLL receive higher weights. The resulting weights $\{w^{\rm (ens)}_m\}$ are normalized to sum to unity and used to form a weighted average of the member predictions,
\[
\hat{\boldsymbol{\theta}}_{\rm ens}
   = \sum_{m} w^{\rm (ens)}_m\,
     \hat{\boldsymbol{\theta}}^{(m)}.
\]
This ensemble prediction is then passed through the previously calibrated likelihood model to obtain the final posterior mean and uncertainty for each map. We refer the reader to Appendix~\ref{app:nll_weights} for further details on the weighting scheme.

\subsection{Temperature Calibration} \label{app:tau}

Let $\{\hat{\boldsymbol{\theta}}_i\}_{i=1}^{N_{\rm val}}$ be the CNN
predictions on the concatenated validation set, and let
$\boldsymbol{\theta}_{g(i)}$ denote the true cosmology of sample $i$,
corresponding to grid index $g(i)$. We have, at each point $g$, a calibrated mean and covariance
$(\bar{\boldsymbol{\mu}}_g, \bar{\mathbf{\Sigma}}_g)$ defining a
Gaussian likelihood model.

For each validation sample we form the residual
\begin{equation}
    \boldsymbol{r}_i
    = \hat{\boldsymbol{\theta}}_i - \bar{\boldsymbol{\mu}}_{g(i)},
\end{equation}
and compute the whitened residual using the Cholesky factor
$\bar{\mathbf{\Sigma}}_{g(i)} = \mathbf{L}_{g(i)} \mathbf{L}_{g(i)}^\top$:
\begin{equation}
    \boldsymbol{z}_i = \mathbf{L}_{g(i)}^{-1}\,\boldsymbol{r}_i,
    \qquad
    q_i = \|\boldsymbol{z}_i\|_2^2.
\end{equation}
For perfectly calibrated Gaussian residuals in $d$ dimensions, the
Mahalanobis distances $q_i$ follow a $\chi^2_d$ distribution with
$\mathbb{E}[q_i] = d$. In practice we allow a free ``target degrees of
freedom'' parameter $p_{\rm DOF}$ and choose a global temperature
$\tau$ such that the empirical mean of the scaled distances matches
$p_{\rm DOF}$:
\begin{equation}
    \frac{1}{N_{\rm val}} \sum_{i=1}^{N_{\rm val}}
    \frac{q_i}{\tau^2}
    \;=\;
    p_{\rm DOF}
    \quad\Longrightarrow\quad
    \tau^2
    =
    \frac{\frac{1}{N_{\rm val}}\sum_i q_i}{p_{\rm DOF}}.
\end{equation}
Finally, we rescale all grid covariances by this temperature factor,bwhich inflates (or deflates) the uncertainties so that the distribution
of whitened residuals on the validation set is consistent with the
target $\chi^2_{p_{\rm DOF}}$ behaviour.

\subsection{Ensemble Weighting by Negative Log-Likelihood}
\label{app:nll_weights}

Suppose we have $M$ independently trained ensemble members, and let
$\hat{\boldsymbol{\theta}}^{(m)}_i$ denote the test prediction of
member $m$ for map $i$, with $i = 1,\dots,N_{\rm test}$. Using the
calibrated grid likelihood
$\mathcal{N}(\bar{\boldsymbol{\mu}}_g,\widetilde{\mathbf{\Sigma}}_g)$
from above and a uniform prior over the cosmology grid, we approximate
the predictive marginal for a single prediction as
\begin{equation}
    p_m\!\bigl(\hat{\boldsymbol{\theta}}^{(m)}_i\bigr)
    \;\approx\;
    \frac{1}{G}
    \sum_{g=1}^{G}
    \mathcal{N}\!\bigl(
        \hat{\boldsymbol{\theta}}^{(m)}_i
        \,;\,
        \bar{\boldsymbol{\mu}}_g,
        \widetilde{\mathbf{\Sigma}}_g
    \bigr),
\end{equation}
where $G$ is the number of grid points. For each member $m$ we then
compute its mean marginal log-likelihood over all test maps
\begin{equation}
    \ell_m
    \;=\;
    \frac{1}{N_{\rm test}}
    \sum_{i=1}^{N_{\rm test}}
    \log p_m\!\bigl(\hat{\boldsymbol{\theta}}^{(m)}_i\bigr),
\end{equation}
and define the corresponding negative log-likelihood
\begin{equation}
    {\rm NLL}_m = -\ell_m.
\end{equation}
Members that assign higher probability to their own predictions (i.e.\ 
lower NLL) are thus preferred. We convert these NLL values into
non-negative ensemble weights via a softmax,
\begin{equation}
    w_m
    =
    \frac{\exp(-{\rm NLL}_m)}
         {\sum_{m'=1}^{M}\exp(-{\rm NLL}_{m'})},
    \qquad
    \sum_{m=1}^{M} w_m = 1.
\end{equation}
The final ensemble prediction for each test map is then the weighted
average of the member predictions in parameter space,
\begin{equation}
    \hat{\boldsymbol{\theta}}^{\rm (ens)}_i
    =
    \sum_{m=1}^{M}
    w_m\,\hat{\boldsymbol{\theta}}^{(m)}_i,
\end{equation}
which is subsequently passed through the calibrated grid likelihood to
obtain the posterior mean and uncertainties.

\section{Inception and InceptionSE Architecture} \label{app:inceptionse}

In Table~\ref{tbl:backbones} we provide a layer-by-layer summary of the Inception and InceptionSE model we used.

\begin{table*}
\centering
\caption{Layer-by-layer summary of the two backbone architectures. The total number of trainable parameters is 422{,}754 for the Inception model and 722{,}610 for the Inception--SE model.}
\label{tbl:backbones}
\renewcommand{\arraystretch}{1.35}
\begin{tabular}{|c|c|c|c|}
\hline
\multicolumn{2}{|c|}{\textbf{Inception}} & \multicolumn{2}{c|}{\textbf{InceptionSE}} \\
\hline
\textbf{Operation} & \textbf{Output size} & \textbf{Operation} & \textbf{Output size} \\
\hline

\begin{tabular}[t]{@{}l@{}}
Conv. ($7\times7$, $s=2$) \\
MaxPool ($2\times2$)
\end{tabular} &
$32\times \tfrac{H}{4} \times \tfrac{W}{4}$ &
\begin{tabular}[t]{@{}l@{}}
Conv. ($7\times7$, $s=2$) \\
MaxPool ($2\times2$)
\end{tabular} &
$48\times \tfrac{H}{4} \times \tfrac{W}{4}$ \\
\hline

\begin{tabular}[t]{@{}l@{}}
\textbf{Inception}: \\
Conv. ($1\times1$) \\
Conv. ($3\times3$)  \\
Conv. ($5\times5$) \\
MaxPool ($3 \times 3$) + Conv. ($1\times1$) \\
MaxPool ($2 \times 2$)
\end{tabular} &
$64\times \tfrac{H}{8}\times \tfrac{W}{8}$ &
\begin{tabular}[t]{@{}l@{}}
\textbf{Inception}: \\
Conv. ($1\times1$) \\
Conv. ($3\times3$)  \\
Conv. ($5\times5$) \\
MaxPool ($3 \times 3$) + Conv. ($1\times1$) \\
\textbf{SE}: $r=8$\\
MaxPool ($2 \times 2$)
\end{tabular} &
$96\times \tfrac{H}{8}\times \tfrac{W}{8}$ \\
\hline

\begin{tabular}[!ht]{@{}l@{}}
\textbf{Inception}: \\
Conv. ($1\times1$) \\
Conv. ($3\times3$)  \\
Conv. ($5\times5$) \\
MaxPool ($3 \times 3$) + Conv. ($1\times1$) \\
MaxPool ($2 \times 2$)
\end{tabular} &
$128\times \tfrac{H}{16}\times \tfrac{W}{16}$ &
\begin{tabular}[t]{@{}l@{}}
\textbf{Inception}: \\
Conv. ($1\times1$) \\
Conv. ($3\times3$)  \\
Conv. ($5\times5$) \\
MaxPool ($3 \times 3$) + Conv. ($1\times1$) \\
\textbf{SE}: $r=8$\\
MaxPool ($2 \times 2$)
\end{tabular} &
$192\times \tfrac{H}{16}\times \tfrac{W}{16}$ \\
\hline

\begin{tabular}[t]{@{}l@{}}
\textbf{Inception}: \\
Conv. ($1\times1$) \\
Conv. ($3\times3$)  \\
Conv. ($5\times5$) \\
MaxPool ($3 \times 3$) + Conv. ($1\times1$) \\
MaxPool ($2 \times 2$)
\end{tabular} &
$256\times \tfrac{H}{16}\times \tfrac{W}{16}$ &
\begin{tabular}[t]{@{}l@{}}
\textbf{Inception}: \\
Conv. ($1\times1$) \\
Conv. ($3\times3$)  \\
Conv. ($5\times5$) \\
MaxPool ($3 \times 3$) + Conv. ($1\times1$) \\
\textbf{SE}: $r=8$\\
MaxPool ($2 \times 2$)
\end{tabular} &
$256\times \tfrac{H}{16}\times \tfrac{W}{16}$ \\
\hline

AdaptiveAvgPool ($1\times1$) &
$\mathbb{R}^{256}$ &
AdaptiveAvgPool ($1\times1$) &
$\mathbb{R}^{256}$ \\
\hline

FC: $256 \rightarrow 128 \rightarrow 2$ &
$\mathbb{R}^{2}$ &
FC: $256 \rightarrow 160 \rightarrow 2$ &
$\mathbb{R}^{2}$ \\
\hline

\end{tabular}
\end{table*}

\section{Scattering Covariances} \label{app:sc}

\subsection{Definition}

The SC statistics are constructed through successive wavelet transforms and modulus operators applied to a field $I$, followed by spatial mean or covariance estimations. We refer to~\cite{Cheng2024} for full details on the construction of these statistics and give a brief overview here. Each oriented wavelet $\psi^{\lambda_{i}}$ is indexed by $\lambda_{i} = (j_{i}, \theta_{i})$, denoting its dyadic scale as well as orientation. We used $6$ dyadic scales (limited by the width of the maps) and $4$ angles. We consider 4 types of coefficients: 

\begin{equation}\label{eq:S1}
    S_{1}^{\lambda_{1}} = \langle | I\star \psi^{\lambda_{1}} | \rangle,
\end{equation}

\begin{equation}\label{eq:S2}
    S_{2}^{\lambda_{1}} = \langle | I\star \psi^{\lambda_{1}} |^{2} \rangle,
\end{equation}

\begin{equation}\label{eq:S3}
    S_{3}^{\lambda_{1},\lambda_{2}} = Cov \left[I \star \psi^{\lambda_{1}}, |I \star \psi^{\lambda_{2}}| \star \psi^{\lambda_{1}} \right],
\end{equation}

\begin{equation}\label{eq:S4}
    S_{4}^{\lambda_{1},\lambda_{2},\lambda_{3}} =  Cov \left[|I \star \psi^{\lambda_{3}}| \star \psi^{\lambda_{1}}, |I \star \psi^{\lambda_{2}}| \star \psi^{\lambda_{1}} \right],
\end{equation}

where $\langle \cdot \rangle$ corresponds to a spatial average and $Cov \left[XY\right] = \langle XY^{*} \rangle - \langle X \rangle \langle Y^{*} \rangle$ is the estimated spatial covariance for two complex fields $X$ and $Y$. We reduced the dimensionality of the  representation by considering isotropic coefficients leading to a dimension of around $600$. The representation could be further reduced by considering only certain scales, coefficients or performing a fourier transform on the scales and selecting only a few harmonics (see~\cite{Cheng2024} for more details). We did not have time to investigate these kind of compressions and leave it for future work.  We computed all SC coefficients using the foscat package $\textbf{foscat}$\footnote{https://github.com/jmdelouis/FOSCAT}, which supports convolutions on rectangular and masked maps.

\subsection{Integration}

To integrate the SC statistics with the outputs of the CNNs with our pipeline, we first applied PCA to reduce the SC feature dimensionality from 630 to 64, which acts as an effective denoising and regularization step. Next, we train a lightweight regression MLP on top of the PCA-compressed SC representation and combine it with the pretrained CNN via
\begin{align}
    \hat{y} &= \hat{y}_\text{cnn} + \alpha \hat{y}_\text{mlp}
\end{align}
where $\alpha$ is a trainable scalar initialized at 0.01. Because the pretrained CNN already achieves strong performance, the SC correction term provides only a slight improvement in score of +0.01, but it does so consistently across experiments.

We trained the combined SC and CNN ensembles for a further 25 epochs. We use the same hyper-parameters for each model within the ensemble with the exception of reducing the base learning rate of the CNN by a factor of 10 to $1\times10^{-5}$. This prevents large, damaging updates away from the already accurate CNN predictions, whilst also allowing small deviations away to better interact with the SC MLP.

\section{Performance over the Parameter Space} \label{app:param-space}

\begin{figure}[!ht]
    \begin{subfigure}[b]{0.49\textwidth}
        \centering
        \includegraphics[width=\textwidth]{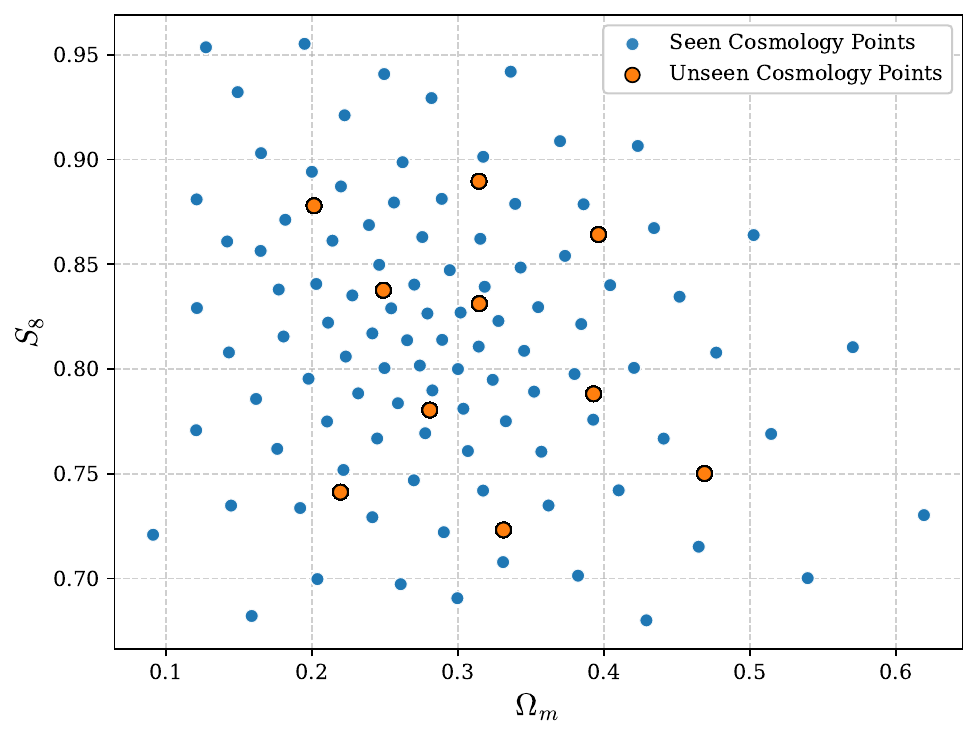}
    \end{subfigure}
    \hfill
    \begin{subfigure}[b]{0.49\textwidth}
         \centering
         \includegraphics[width=\textwidth]{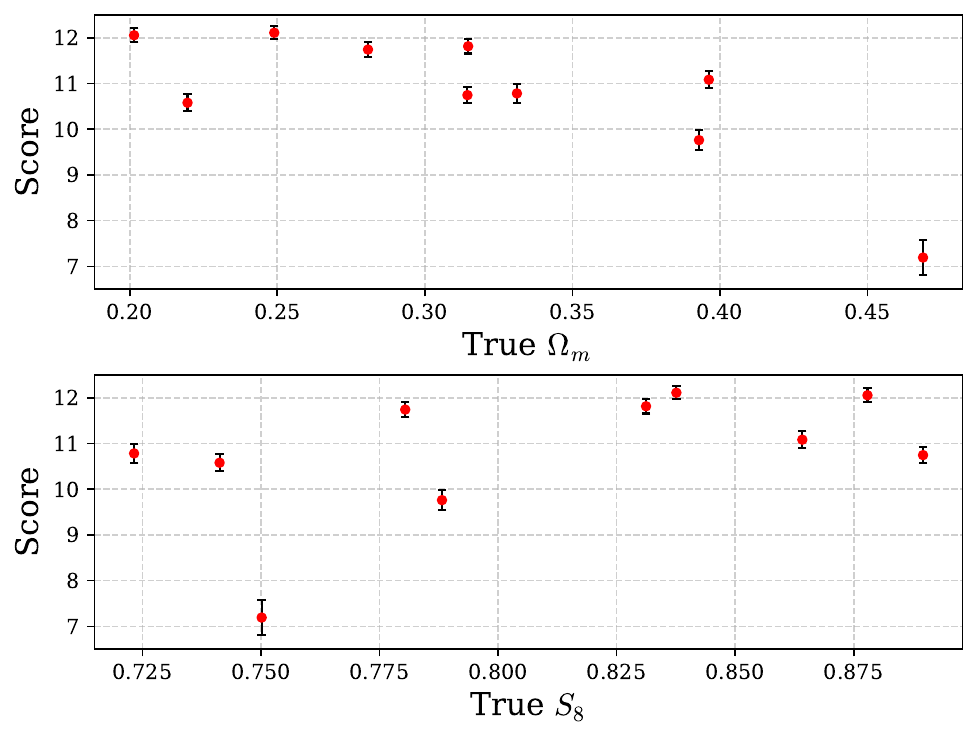}
     \end{subfigure}
     \caption{\textbf{Left:} The cosmologies seen during training, validation, and testing, and the unseen cosmologies only seen during testing. \textbf{Right:} The mean score of our inference pipeline for each unseen cosmology with the associated standard error on the mean. The poor performance at one unseen cosmology is driven by the scarcity of the training data in that region.}
     \label{fig:cosmology_grid}
\end{figure}

To provide a comprehensive assessment of our agent-driven discovery process, we also report the performance in a different regime. The score optimized during development, and reported in Table \ref{tbl:final_results}, primarily reflects test cases whose cosmology labels were represented in the training set. In the full test dataset, however, a subset of cosmologies lies off this training grid. Looking at Figure \ref{fig:cosmology_grid}, our inference pipeline performs well when within a region with a high density of training cosmologies. In contrast, our pipeline performs much worse at $\theta = (0.47, 0.75)$. This decline is attributed to data scarcity within that specific region of the parameter space. A denser simulation suite would likely improve the generalization of the final inference pipeline across the full parameter space. This problem affected all submissions and was not unique to our solution.

\newpage

\section{Example Prompt} \label{app:prompt}

An example prompt with the structure we used during the autonomous exploration by \texttt{Cmbagent}. For human intervention, we modify this further with our own insights for example by emphasizing that ResNet18 or smaller models should be used.

\begin{figure}[!h]
\centering
\fbox{
\begin{minipage}{0.95\linewidth}
\textbf{Find and train a neural network that maximises the score. Best model will achieve above 11.}

\textbf{Previous run insights:} \\
Baseline Simple\_CNN Score:  $\sim8.2-8.5$ \\
Single ResNet18 Score: 8.91 \\
... \\
\\
\textbf{Key findings might be important to improve the score:} \\
Data Augmentation: … \\
Ensembling: … \\
\\
\textbf{However, fundamental shift in modelling approach is required to improve further, e.g.:} \\
Architecture modification: … \\
Loss Function: … \\

\textbf{Here is the example code to load the training images:} \\
(example codes)\\
\\
\textbf{Hardware constraints:} \\
We are running on an NVIDIA RTX PRO 6000 Blackwell Workstation Edition with 96GB RAM
\end{minipage}
}
\caption{Example prompt}
\label{fig:prompt_example}
\end{figure}

\end{document}